\documentclass[11pt]{article}

\usepackage[preprint]{acl}

\usepackage{times}
\usepackage{latexsym}
\usepackage[T1]{fontenc}
\usepackage[utf8]{inputenc}
\usepackage{microtype}
\usepackage{inconsolata}
\usepackage{graphicx}
\usepackage{booktabs}
\usepackage{amsmath,amssymb,amsfonts,amsthm}
\usepackage{bm}
\usepackage{bbm}
\usepackage{multirow}
\usepackage{xcolor}
\usepackage{colortbl}
\usepackage{array}
\usepackage{tabularx}
\usepackage{hyperref}
\usepackage{enumitem}
\usepackage[ruled,vlined]{algorithm2e}
\SetKwInput{KwInput}{Input}
\SetKwInput{KwOutput}{Output}

\newtheorem{proposition}{Proposition}
\newtheorem{theorem}{Theorem}
\newtheorem{lemma}{Lemma}
\newtheorem{corollary}{Corollary}

\newtheorem{assumption}{Assumption}

\makeatletter
\renewcommand{\paragraph}{\@startsection{paragraph}{4}{\z@}{0.85ex plus 0.25ex minus .1ex}{-1em}{\normalsize\bfseries}}
\makeatother

\definecolor{oursbg}{RGB}{253,238,240}

\newcommand{\method}{\textsc{Prism}}
\newcommand{\E}{\mathbb{E}}
\newcommand{\R}{\mathbb{R}}
\newcommand{\ip}[2]{\langle #1,#2\rangle}
\newcommand{\ipM}[2]{\langle #1,#2\rangle_{M}}
\newcommand{\nrmM}[1]{\|#1\|_{M}}
\newcommand{\frob}{\mathrm{F}}

\title{\method{}: Validation-Free Online Batch Selection for LLM Fine-Tuning\\via Preconditioned Conic Ghost Matching}

\author{
  \textbf{Lin Tang\textsuperscript{1}},
  Wei Zhang\textsuperscript{1},
  Jing Li\textsuperscript{1},
  Hongyu Chen\textsuperscript{1},
  Ming Zhao\textsuperscript{2},
  Yuxuan Wang\textsuperscript{2}
  \\
  \\
  \textsuperscript{1}Sichuan University, Chengdu, China \\
  \textsuperscript{2}University of Electronic Science and Technology of China, Chengdu, China
}

\begin{document}
\maketitle

\begin{abstract}
Online batch selection lowers the cost of fine-tuning large language models (LLMs) by replacing each candidate batch with a smaller informative subset. Existing selectors typically depend on a held-out validation or proxy set, or else rely on Euclidean facility-location objectives over unpreconditioned gradients whose diameter-shifted similarities are distorted by heavy-tailed gradient norms and cross-layer anisotropy. We introduce \method{} (\emph{Preconditioned Ray-Isotropic Submodular Matching}), a validation-free online batch selector that aligns candidate gradients in a factored second-moment metric and casts subset construction as \emph{conic facility location} over positive ray projections. Computing the required pairwise geometry through factored preconditioned ghost inner products avoids materializing per-example parameter gradients, while the conic objective remains normalized, nonnegative, and monotone submodular---guaranteeing a $(1-1/e)$ greedy approximation whose additive error scales with total batch gradient energy rather than the worst-case outlier diameter. Assigning each candidate to its best-aligned selected ray induces closed-form radial projection weights that automatically shrink orthogonal or noisy examples toward zero. We derive a nonconvex convergence bound showing that optimization error is governed directly by the conic representation cost. Across four LLM fine-tuning configurations spanning 1.5B to 8B parameters, \method{} outperforms full-batch training, Euclidean facility location, and zeroth-order selectors, and rivals or surpasses validation-guided baselines without held-out data.
\end{abstract}

\section{Introduction}
\label{sec:intro}

Fine-tuning large language models (LLMs) on instruction-following and reasoning mixtures is compute-intensive, and examples within a candidate batch vary sharply in informativeness and gradient quality~\citep{albalak2024survey,hu2022lora,zou2025utility}. Online batch selection addresses this by scoring a candidate pool $\mathcal{B}_t$ of $b$ examples at each step and updating parameters on a selected subset $\mathcal{S}_t$ of size $k < b$~\citep{loshchilov2015online,katharopoulos2018not}. Because exhaustive subset evaluation is combinatorially intractable, practical selectors construct $\mathcal{S}_t$ greedily; when the selection objective is monotone submodular, greedy construction inherits the classical $(1-1/e)$ approximation guarantee~\citep{nemhauser1978analysis}, whereas non-submodular heuristics carry no structural certification.

Recent LLM selectors fall into two main families. Validation-guided methods such as GREATS~\citep{wang2024greats}, OPUS~\citep{wang2026opus}, BLADE~\citep{wang2026blade}, and Filter-then-Weight~\citep{wang2026filter} score candidates via alignment with a held-out validation or proxy set, leveraging ghost inner products or optimizer-aware projections. While effective when a clean, task-matched validation set is known beforehand, such supervision is unavailable in continual, multi-task, or open-ended instruction tuning~\citep{peng2024energy,zhang2026coda}, and Taylor-expanded validation surrogates need not be submodular. Validation-free methods eliminate external data by summarizing the batch's own geometry---either via utility-diversity scores~\citep{zou2025utility,zou2026slap} or gradient-matching coresets such as CoLM~\citep{nguyen2025minibatch}, GRACE~\citep{tang2026grace}, and Euclidean facility location~\citep{mirzasoleiman2020coresets,killamsetty2021gradmatch}.

\begin{figure*}[t]
    \centering
    \includegraphics[width=\textwidth]{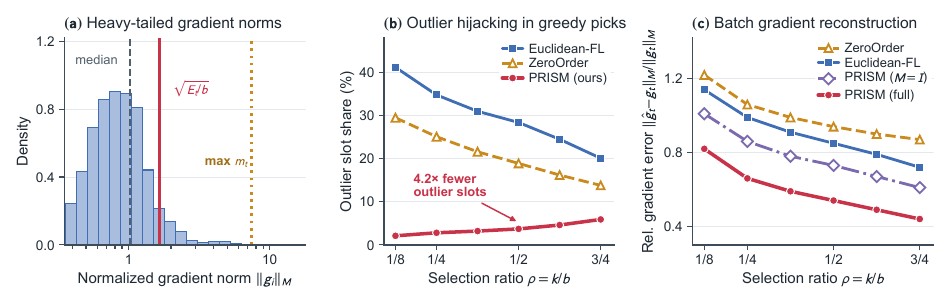}
    \caption{\textbf{Heavy-tailed norms distort Euclidean facility location, while conic matching suppresses outliers.} (a) Per-example gradient norms on Tulu-2 (Llama-3.1-8B) exhibit a heavy tail: the worst-case Euclidean diameter $m_t$ sits far above the batch RMS scale $\sqrt{E_t/b}$. (b) Because Euclidean similarity $m_t - \|g_i-g_j\|$ rewards covering isolated large-norm points, Euclidean-FL spends $28.4\%$ of its greedy slots on outliers at $\rho{=}1/2$, whereas \method{} cuts outlier slots to $3.7\%$. (c) Across selection ratios $\rho$, \method{} achieves the lowest normalized gradient reconstruction error.}
    \label{fig:motivation}
\end{figure*}

However, standard Euclidean facility location over raw gradients suffers from two structural weaknesses in LLM fine-tuning. First, LLM gradients exhibit heavy-tailed norms across sequence lengths and noisy responses~\citep{zhang2020why,yu2026sign}, and transformer layers exhibit strong scale anisotropy~\citep{kim2026form,fan2026optiselect}. Converting Euclidean distances $d_{ij} = \|g_i - g_j\|_2$ into nonnegative similarities via $s_{ij} = m_t - d_{ij}$ anchors the entire objective to the batch's worst-case outlier diameter $m_t = \max_{i,j} d_{ij}$. As Figure~\ref{fig:motivation}(a--b) illustrates, isolated high-norm gradients inflate $m_t$, hijack greedy facility slots, and force uniform integer weights $\gamma_j = |C_j|$ onto misaligned neighbors. Second, evaluating raw Euclidean distances ignores the diagonal curvature geometry of adaptive optimizers such as AdamW~\citep{loshchilov2019decoupled,ding2026inrun}.

We propose \method{} (\emph{Preconditioned Ray-Isotropic Submodular Matching}), a validation-free online batch selector built around two coupled geometric ideas. First, instead of Euclidean balls in raw parameter space, \method{} operates in a \emph{factored preconditioned metric} $M_t$ using row- and column-scaled ghost inner products, faithfully capturing AdamW's layerwise curvature without materializing per-example gradients. Second, \method{} replaces diameter-shifted Euclidean distances with \emph{conic (ray) similarity}, $s_{ij}^{\mathrm{cone}} = [\max(0, \langle \tilde{g}_i, \hat{g}_j \rangle_M)]^2$, measuring the squared nonnegative projection of candidate $i$ onto the unit ray $\hat{g}_j$ of representative $j$. The resulting conic facility-location objective is naturally nonnegative, normalized, and monotone submodular without any diameter shift $m_t$. Assigning each candidate to its best-aligned ray yields \emph{closed-form radial projection weights} $\gamma_{t,j} = \sum_{i \in C_j} \max(0, \langle \tilde{g}_i, \tilde{g}_j\rangle_M)/\|\tilde{g}_j\|_M^2$ that preserve directional signal while automatically assigning zero weight to orthogonal or noisy outliers (Figure~\ref{fig:motivation}c).

We summarize our contributions as follows:
\begin{itemize}[leftmargin=*,itemsep=1pt,topsep=1pt]
\item \textbf{Conic ghost matching.} We formulate validation-free online batch selection as conic facility location over factored preconditioned ghost inner products with closed-form radial weights (\S\ref{sec:method}).
\item \textbf{Tighter combinatorial and optimization theory.} We prove that \method{} keeps the $(1-1/e)$ greedy guarantee while replacing the outlier diameter term $bm_t/e$ with total batch energy $E_t/e$, and derive nonconvex and PL convergence bounds governed by conic cost (\S\ref{sec:theory}, Appendix~\ref{app:extra_theory}).
\item \textbf{Empirical gains and mechanism evidence.} Across four LLM fine-tuning tasks (1.5B--8B), \method{} outperforms full-batch training, Euclidean facility location, and zeroth-order selection, rivals validation-guided baselines without held-out data, and rejects injected label noise (\S\ref{sec:experiments}).
\end{itemize}

\section{Related Work}
\label{sec:related}

\paragraph{Online batch selection for LLMs.}
Early online selectors prioritize high-loss or large-gradient examples~\citep{loshchilov2015online,katharopoulos2018not}. Subsequent reference-based methods score candidates against an auxiliary model trained on held-out data~\citep{mindermann2022prioritized,evans2024data}, culminating in bi-level and adaptive reference selectors such as BLADE~\citep{wang2026blade}. For LLM training, GREATS~\citep{wang2024greats} approximates one-step validation-loss reduction via ghost inner products, while OPUS~\citep{wang2026opus}, Filter-then-Weight~\citep{wang2026filter}, and Gaussian-process online selection~\citep{wang2026gaia} incorporate optimizer preconditioning or proxy alignment. Recent work also investigates on-policy attribution~\citep{wang2026drift}, learning-zone selection in RL post-training~\citep{cui2026learning}, and implicit alignment effects of online filtering~\citep{zeng2026online}. Unlike these validation- or proxy-directed selectors, \method{} requires no held-out data and operates purely on intra-batch conic gradient geometry.

\paragraph{Gradient coresets and submodular selection.}
Offline and epoch-level gradient coresets such as CRAIG~\citep{mirzasoleiman2020coresets}, GRAD-MATCH~\citep{killamsetty2021gradmatch}, TAGCOS~\citep{zhang2025tagcos}, and InfoMax~\citep{tan2025infomax} match the full training gradient via facility location, orthogonal matching pursuit, or quadratic information maximization, while newer 2025--2026 pruning pipelines explore iterative distillation, semantic manifolds, and dual-budget hidden-state routing~\citep{lin2025lead,hu2025donod,nikdan2025efficient,sun2026rethinking,qin2026hidden}. At the per-batch LLM level, CoLM~\citep{nguyen2025minibatch} applies mini-batch coresets over source-clustered zeroth-order gradient estimates, GRACE~\citep{tang2026grace} maintains dynamic gradient coresets, SLAP~\citep{zou2026slap} uses stratified Hessian-approximated pruning, and UDS~\citep{zou2025utility} scores logits nuclear norm alongside embedding diversity. \method{} shares the validation-free coreset ethos but replaces Euclidean distance matching and zeroth-order sketches with exact factored-metric conic facility location and radial projection weights.

\paragraph{Preconditioned geometry and gradient noise.}
Ghost gradient computations scale per-example inner products to large transformers~\citep{li2022large,wang2024greats}, enabling in-run attribution~\citep{wang2025data,ding2026inrun} and scalable attribution libraries~\citep{liu2026dattri}. At the same time, recent studies show that raw gradient inner products are skewed by layer norm anisotropy, heavy-tailed noise, and optimizer dynamics~\citep{zhang2020why,kim2026form,fan2026optiselect,yu2026sign}, and that gradient signal-to-noise structure governs instruction-tuning quality~\citep{yuan2026uncertainty,han2026what,necula2026highquality}. \method{} bridges these findings by embedding a Kronecker-factored second-moment metric directly inside ghost inner products and using conic projection to separate coherent directional signal from norm outliers.

\section{Preliminaries and Problem Formulation}
\label{sec:prelim}

\paragraph{Online batch selection.}
Let $\mathcal{D}=\{z_i\}_{i=1}^n$ be a fine-tuning dataset and $\ell(w,z)$ the per-example loss of parameters $w \in \R^p$, defining empirical risk $f(w) = \frac{1}{n}\sum_{z\in\mathcal{D}}\ell(w,z)$. At iteration $t$, we sample a candidate batch $\mathcal{B}_t = \{z_1,\dots,z_b\} \subseteq \mathcal{D}$ with per-example gradients $g_{t,i} = \nabla_w \ell(w_t,z_i)$ and full-batch mean $g_t = \frac{1}{b}\sum_{i=1}^b g_{t,i}$. Our goal is to select a smaller subset $\mathcal{S}_t \subseteq \mathcal{B}_t$ of cardinality $|\mathcal{S}_t| = k < b$ and nonnegative weights $\{\gamma_{t,j}\}_{j\in\mathcal{S}_t}$ so that the weighted subset gradient $\hat{g}_t = \frac{1}{b}\sum_{j\in\mathcal{S}_t} \gamma_{t,j} g_{t,j}$ faithfully tracks $g_t$ in a positive-definite metric $M_t \succ 0$:
\begin{equation}
\min_{\substack{\mathcal{S}\subseteq\mathcal{B}_t,\,|\mathcal{S}|=k \\ \gamma_j \ge 0}} \;\Big\|\, g_t - \frac{1}{b}\sum_{j\in\mathcal{S}}\gamma_j g_{t,j} \,\Big\|_{M_t}^2,
\label{eq:match_obj}
\end{equation}
where $\|u\|_M^2 = u^\top M u$ and $\ipM{u}{v} = u^\top M v$. Unlike classical formulations that enforce $\sum_{j\in\mathcal{S}}\gamma_j = b$~\citep{mirzasoleiman2020coresets,killamsetty2021gradmatch}, leaving $\gamma_j$ unconstrained in magnitude allows radial shrinkage of poorly aligned batch regions while retaining a single weighted backward pass over $\mathcal{S}_t$.

\paragraph{Ghost inner products.}
Materializing $G_t \in \R^{b\times p}$ explicitly is prohibitive for LLMs. For any linear layer $y = Wx$ with input activation $x_i \in \R^{d_{\mathrm{in}}}$ and backpropagated signal $\theta_i = \nabla_{y_i}\ell_i \in \R^{d_{\mathrm{out}}}$, the parameter gradient is $\nabla_W \ell_i = \theta_i x_i^\top$, so the Frobenius inner product factorizes as~\citep{li2022large,wang2024greats}
\begin{equation}
\ip{\nabla_W \ell_i}{\nabla_W \ell_j}_{\frob} = \ip{\theta_i}{\theta_j}\,\ip{x_i}{x_j}.
\label{eq:ghost}
\end{equation}
Summing~\eqref{eq:ghost} across all trainable linear and LoRA modules yields the exact $b\times b$ Gram matrix $K_t$ using only $O(b^2)$ activation and error-signal products, without storing $p$-dimensional per-example gradients.

\begin{figure*}[t]
    \centering
    \includegraphics[width=\textwidth]{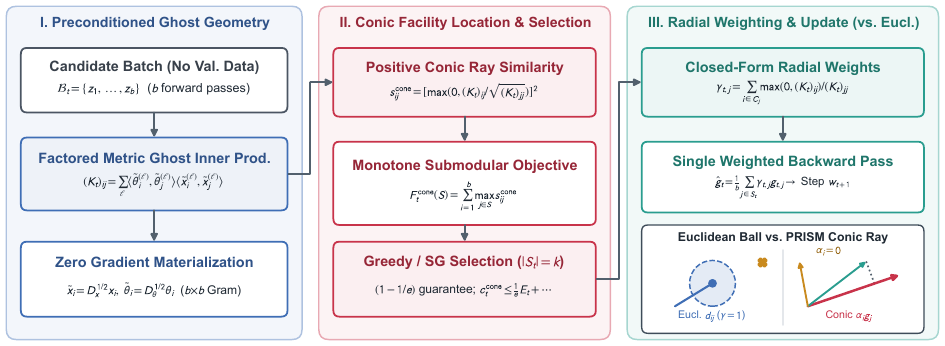}
    \caption{\textbf{Overview of \method{}.} \textbf{I.} Forward activations $x_i$ and backward signals $\theta_i$ are rescaled by factored second-moment statistics $D_x^{1/2}, D_\theta^{1/2}$ to form the preconditioned $b{\times}b$ Gram matrix $K_t$ via ghost inner products with zero per-example gradient materialization. \textbf{II.} Pairwise entries induce nonnegative conic ray similarities $s_{ij}^{\mathrm{cone}}$, defining a monotone submodular objective $F_t^{\mathrm{cone}}$ maximized greedily (or via stochastic greedy). \textbf{III.} Each candidate projects onto its closest ray to yield closed-form radial weights $\gamma_{t,j}$ (automatically shrinking orthogonal outliers to $\alpha_i{=}0$), followed by a single weighted backward pass on $\mathcal{S}_t$.}
    \label{fig:architecture}
\end{figure*}

\paragraph{Why Euclidean facility location struggles.}
Standard gradient-space facility location converts pairwise distances $d_{ij} = \|g_{t,i}-g_{t,j}\|_2$ into similarities $s_{ij} = m_t - d_{ij}$ with $m_t = \max_{u,v\in[b]} d_{uv}$ and assigns integer cluster weights $\gamma_j = |C_j|$~\citep{mirzasoleiman2020coresets}. When gradient norms are heavy-tailed, a single large-norm outlier forces $m_t \gg \mathrm{median}_i \|g_{t,i}\|$, inflating the $(1/e)b m_t$ gap in the greedy approximation bound and biasing selection toward isolated extremes. Moreover, integer counts $\gamma_j = |C_j|$ force a representative $g_j$ to carry full unit weight even for outliers that point orthogonally or opposite to $g_j$.

\section{Method: \method{}}
\label{sec:method}

Figure~\ref{fig:architecture} illustrates the three stages of \method{}, which eliminates both diameter distortion and scale anisotropy while preserving $O(b^2)$ ghost efficiency.

\subsection{Factored Preconditioned Ghost Metric}
\label{sec:method-metric}

To capture adaptive-optimizer geometry without breaking the rank-1 factorization~\eqref{eq:ghost}, we equip each trainable matrix $W^{(\ell)} \in \R^{d_{\mathrm{out}}\times d_{\mathrm{in}}}$ (or LoRA factors $A^{(\ell)},B^{(\ell)}$) with a \emph{Kronecker-factored diagonal metric} $M_t^{(\ell)} = D_{\theta,t}^{(\ell)} \otimes D_{x,t}^{(\ell)}$ built from running row- and column-wise second-moment estimates:
\begin{equation}
\begin{aligned}
(D_{\theta,t}^{(\ell)})_{r} &= \big(\sqrt{\hat{v}_{\theta,t-1,r}^{(\ell)}} + \epsilon\big)^{-1},\\
(D_{x,t}^{(\ell)})_{c} &= \big(\sqrt{\hat{v}_{x,t-1,c}^{(\ell)}} + \epsilon\big)^{-1},
\end{aligned}
\label{eq:factored_metric}
\end{equation}
where $\hat{v}_{\theta,t-1}^{(\ell)}$ and $\hat{v}_{x,t-1}^{(\ell)}$ are exponential moving averages of row and column gradient squares from prior steps (fixed before $\mathcal{B}_t$ is drawn, and updated once per step from the final selected update). Under $M_t = \bigoplus_\ell M_t^{(\ell)}$, the preconditioned inner product factorizes \emph{exactly} into rescaled ghost activations $\tilde{x}_i^{(\ell)} = (D_{x,t}^{(\ell)})^{1/2} x_i^{(\ell)}$ and signals $\tilde{\theta}_i^{(\ell)} = (D_{\theta,t}^{(\ell)})^{1/2} \theta_i^{(\ell)}$:
\begin{equation}
(K_t)_{ij} = \ipM{g_{t,i}}{g_{t,j}} = \sum_{\ell} \ip{\tilde{\theta}_i^{(\ell)}}{\tilde{\theta}_j^{(\ell)}}\,\ip{\tilde{x}_i^{(\ell)}}{\tilde{x}_j^{(\ell)}}.
\label{eq:precond_ghost}
\end{equation}
For LoRA adapters, the same factorization applies to $A^{(\ell)}$ and $B^{(\ell)}$ (Appendix~\ref{app:impl}). Thus, \method{} computes the preconditioned Gram matrix $K_t \in \R^{b\times b}$ with $O(d_{\mathrm{in}}+d_{\mathrm{out}})$ metric state per layer and zero materialization of per-example parameter gradients.

\subsection{Conic Facility Location and Radial Weighting}
\label{sec:method-conic}

Let $\tilde{g}_{t,i} = M_t^{1/2} g_{t,i}$ denote the preconditioned gradient of candidate $i$, whose squared norm is $(K_t)_{ii}$. For any candidate $i \in [b]$ and potential representative $j \in [b]$ with $(K_t)_{jj} > 0$, let $\hat{g}_{t,j} = \tilde{g}_{t,j} / \sqrt{(K_t)_{jj}}$ be the unit ray along $\tilde{g}_{t,j}$. The optimal nonnegative radial projection of $\tilde{g}_{t,i}$ onto the ray $\{\alpha \hat{g}_{t,j} : \alpha \ge 0\}$ is $p_{ij} = \alpha_{ij}\tilde{g}_{t,j}$ with $\alpha_{ij} = \max(0, (K_t)_{ij}) / (K_t)_{jj}$, which attains squared reconstruction error
\begin{equation}
\min_{\alpha \ge 0} \|\tilde{g}_{t,i} - \alpha \hat{g}_{t,j}\|_2^2 = (K_t)_{ii} - s_{t,ij}^{\mathrm{cone}},
\label{eq:ray_err}
\end{equation}
where the \emph{conic ray similarity} is defined from $K_t$:
\begin{equation}
\begin{aligned}
s_{t,ij}^{\mathrm{cone}} &:= \frac{\big[\max\!\big(0,\,(K_t)_{ij}\big)\big]^2}{(K_t)_{jj} + \delta} \\
&= (K_t)_{ii}\cos_+^2(\tilde{g}_{t,i},\tilde{g}_{t,j}),
\end{aligned}
\label{eq:conic_sim}
\end{equation}
with $\delta{=}10^{-12}$. Summing~\eqref{eq:ray_err} over $i \in [b]$ yields the \emph{conic representation cost}
\begin{equation}
\begin{aligned}
c_t^{\mathrm{cone}}(\mathcal{S}) &:= \sum_{i=1}^b \min_{j\in\mathcal{S}} \big[(K_t)_{ii} - s_{t,ij}^{\mathrm{cone}}\big] \\
&= E_t - F_t^{\mathrm{cone}}(\mathcal{S}),
\end{aligned}
\label{eq:conic_cost}
\end{equation}
where $E_t := \sum_{i=1}^b (K_t)_{ii}$ is the total preconditioned batch energy and
\begin{equation}
F_t^{\mathrm{cone}}(\mathcal{S}) := \sum_{i=1}^b \max_{j\in\mathcal{S}} s_{t,ij}^{\mathrm{cone}}, \qquad F_t^{\mathrm{cone}}(\emptyset) := 0,
\label{eq:conic_fl}
\end{equation}
is our \emph{conic facility-location objective}. Crucially, $s_{t,ij}^{\mathrm{cone}} \ge 0$ holds automatically without subtracting a worst-case diameter $m_t$, and $c_t^{\mathrm{cone}}$ measures squared residual energy orthogonal or opposite to the selected rays.

\paragraph{Closed-form radial projection weights.}
Once $\mathcal{S}_t$ of size $k$ is selected, each candidate $i \in [b]$ is assigned to its best-aligned representative ray $\sigma_{\mathcal{S}_t}(i) \in \arg\max_{j\in\mathcal{S}_t} s_{t,ij}^{\mathrm{cone}}$, partitioning $[b]$ into cells $C_j = \{i \in [b] : \sigma_{\mathcal{S}_t}(i) = j\}$. Rather than assigning integer weight $|C_j|$, each representative aggregates the exact radial projections of its cell:
\begin{equation}
\gamma_{t,j} := \sum_{i \in C_j} \frac{\max\!\big(0,\,(K_t)_{ij}\big)}{(K_t)_{jj} + \delta} \ge 0.
\label{eq:radial_weights}
\end{equation}
Notice two key properties of~\eqref{eq:radial_weights}. First, if candidate $i$ is a corrupted or high-norm outlier nearly orthogonal or negatively aligned with every selected ray, its contribution $\alpha_{i,\sigma(i)} \approx 0$, so it exerts virtually zero pull on the update. Second, by the Pythagorean identity on each ray projection, $\frac{1}{b}\sum_{j\in\mathcal{S}_t}\gamma_{t,j}\tilde{g}_{t,j} = \frac{1}{b}\sum_{i=1}^b p_{i,\sigma(i)}$, giving an exact upper bound on~\eqref{eq:match_obj} (Proposition~\ref{prop:radial_bound}).

\begin{algorithm}[t]
\footnotesize
\DontPrintSemicolon
\SetAlgoLined
\caption{\method{}: Preconditioned Conic Ghost Selection}
\label{alg:prism}
\KwInput{parameters $w_t$; batch $\mathcal{B}_t=\{z_i\}_{i=1}^b$; size $k$; metric $M_t$}
Compute $(K_t)_{ij}$ via~\eqref{eq:precond_ghost} with zero gradient materialization\;
$s_{t,ij}^{\mathrm{cone}} \leftarrow [\max(0, (K_t)_{ij})]^2 / ((K_t)_{jj}+\delta)$;\quad $\mathcal{S} \leftarrow \emptyset$;\quad $u_i \leftarrow 0$\;
\For{$r = 1, \dots, k$}{
  $j^\star \leftarrow \arg\max_{j \in [b]\setminus \mathcal{S}} \sum_{i=1}^b \max(0,\, s_{t,ij}^{\mathrm{cone}} - u_i)$\;
  $\mathcal{S} \leftarrow \mathcal{S} \cup \{j^\star\}$;\quad $u_i \leftarrow \max(u_i,\, s_{t,ij^\star}^{\mathrm{cone}})\ \forall i\in[b]$\;
}
$\sigma(i) \leftarrow \arg\max_{j\in\mathcal{S}} s_{t,ij}^{\mathrm{cone}}$;\quad $\gamma_{t,j} \leftarrow \sum_{i:\sigma(i)=j} \frac{\max(0, (K_t)_{ij})}{(K_t)_{jj}+\delta}$\;
$\hat{g}_t \leftarrow \nabla_w \big[\frac{1}{b}\sum_{j\in\mathcal{S}} \gamma_{t,j}\ell(w_t,z_j)\big]$ via one backward pass on $\mathcal{S}$\;
Update $w_{t+1}$ with $\hat{g}_t$ and update factored metric $M_{t+1}$\;
\end{algorithm}

\subsection{Implementation and Stochastic-Greedy Acceleration}
\label{sec:method-impl}

Algorithm~\ref{alg:prism} summarizes one step of \method{}. Maintaining the running coverage vector $u_i = \max_{j\in\mathcal{S}} s_{t,ij}^{\mathrm{cone}}$ evaluates each greedy step in $O(b^2)$ scalar operations ($O(kb^2)$ total, sub-millisecond for $b \le 64$). For larger candidate pools ($b \ge 64$), we also support a stochastic-greedy variant (\method{}-SG, Appendix~\ref{app:alg}) that samples $q = \lceil (b/k)\log(1/\varepsilon)\rceil$ unselected candidates per step~\citep{mirzasoleiman2015lazier}, cutting selection complexity to $O(b^2 \log(1/\varepsilon))$ while preserving a $(1-1/e-\varepsilon)$ guarantee. Once $\mathcal{S}_t$ and $\{\gamma_{t,j}\}$ are determined, we compute $\hat{g}_t$ via one standard backward pass over the $k$ selected examples with loss multipliers $\gamma_{t,j}/b$.

\begin{table*}[t]
\centering
\footnotesize
\setlength{\tabcolsep}{2.3pt}
\begin{tabular}{l c cccc c}
\toprule
\textbf{Method} & \textbf{Ext.} & \textbf{Qwen2.5-1.5B} & \textbf{Llama-3.1-8B} & \textbf{Mistral-7B} & \textbf{Qwen2.5-7B} & \textbf{Win} \\
 & \textbf{Data} & \textbf{MMLU PPL ($\downarrow$)} & \textbf{BBH EM ($\uparrow$)} & \textbf{TyDiQA F1 ($\uparrow$)} & \textbf{GSM8K Acc ($\uparrow$)} & \textbf{vs.\ Reg.} \\
\midrule
Regular (full $b$) & No & $5.84 \pm 0.41$ & $54.2 \pm 0.7$ & $0.518 \pm 0.011$ & $74.6 \pm 0.6$ & --- \\
Random (half $k$) & No & $6.42 \pm 0.53$ & $51.8 \pm 0.9$ & $0.489 \pm 0.014$ & $71.9 \pm 0.8$ & 0/4 \\
\midrule
GREATS~\citep{wang2024greats} & 5 val. & $\mathbf{4.62 \pm 0.18}$ & $56.4 \pm 0.5$ & $\mathbf{0.654 \pm 0.008}$ & $76.8 \pm 0.5$ & 4/4 \\
OPUS~\citep{wang2026opus} & 5 proxy & $4.79 \pm 0.21$ & $56.9 \pm 0.6$ & $0.584 \pm 0.009$ & $77.1 \pm 0.4$ & 4/4 \\
Filter-then-Weight~\citep{wang2026filter} & 5 proxy & $4.95 \pm 0.24$ & $56.1 \pm 0.6$ & $0.597 \pm 0.010$ & $76.5 \pm 0.5$ & 4/4 \\
\midrule
Euclidean-FL & No & $5.61 \pm 0.36$ & $55.3 \pm 0.6$ & $0.602 \pm 0.013$ & $75.8 \pm 0.6$ & 4/4 \\
ZeroOrder (CoLM-style) & No & $5.92 \pm 0.44$ & $54.6 \pm 0.8$ & $0.579 \pm 0.015$ & $75.1 \pm 0.7$ & 3/4 \\
\rowcolor{oursbg}
\textbf{\method{} (ours)} & No & $5.14 \pm 0.27$ & $\mathbf{57.6 \pm 0.4}$ & $0.631 \pm 0.009$ & $\mathbf{77.9 \pm 0.4}$ & \textbf{4/4} \\
\bottomrule
\end{tabular}
\caption{\textbf{Main results across four LLM fine-tuning configurations} (mean $\pm$ std over 3 seeds; MMLU averages 2 subjects $\times$ 3 seeds, $n{=}6$). Selectors keep $k{=}8$ of $b{=}16$ candidates; Regular trains on all $16$. Best per column in \textbf{bold}. Without held-out validation data, \method{} beats all validation-free baselines on every task and leads overall on BBH and GSM8K.}
\label{tab:main}
\end{table*}

\section{Theoretical Analysis}
\label{sec:theory}

We now establish the combinatorial and optimization guarantees of \method{}; complete proofs and auxiliary results appear in Appendices~\ref{app:proofs} and~\ref{app:extra_theory}.

\begin{proposition}[Monotone submodularity \& energy-relative bound]
\label{prop:submod}
The conic facility-location objective $F_t^{\mathrm{cone}}(\mathcal{S})$ in~\eqref{eq:conic_fl} is normalized ($F_t^{\mathrm{cone}}(\emptyset)=0$), nonnegative, monotone nondecreasing, and submodular. Consequently, greedy maximization returns $\mathcal{S}_G$ of size $k$ satisfying $F_t^{\mathrm{cone}}(\mathcal{S}_G) \ge (1-1/e) F_t^{\mathrm{cone}}(\mathcal{S}^\star)$, where $\mathcal{S}^\star$ is an optimal size-$k$ subset. Moreover, its conic representation cost satisfies
\begin{equation}
c_t^{\mathrm{cone}}(\mathcal{S}_G) \le \frac{1}{e} E_t + \left(1 - \frac{1}{e}\right) c_t^{\mathrm{cone}}(\mathcal{S}^\star).
\label{eq:energy_bound}
\end{equation}
\end{proposition}

\paragraph{Significance of~\eqref{eq:energy_bound}.}
For standard Euclidean facility location, the additive error term is $(1/e)\,b\,m_t$~\citep{mirzasoleiman2020coresets}, which scales with the single largest pairwise distance $m_t = \max_{i,j}\|g_i - g_j\|$ in the batch. In contrast,~\eqref{eq:energy_bound} replaces $b\,m_t$ with the total gradient energy $E_t = \sum_{i=1}^b \|g_{t,i}\|_{M_t}^2$, which averages over the batch and prevents one high-norm outlier from dominating the approximation slack.

\begin{proposition}[Radial reconstruction bound \& outlier damping]
\label{prop:radial_bound}
Let $\hat{g}_t = \frac{1}{b}\sum_{j\in\mathcal{S}_t}\gamma_{t,j}g_{t,j}$ use the closed-form radial weights~\eqref{eq:radial_weights}. Then:
(i) the preconditioned gradient-matching error obeys
\begin{equation}
\nrmM{g_t - \hat{g}_t}^2 \le \frac{1}{b}\,c_t^{\mathrm{cone}}(\mathcal{S}_t) =: \rho_t^2;
\label{eq:recon_bound}
\end{equation}
(ii) if an example $z_i$ satisfies $\ipM{g_{t,i}}{g_{t,j}} \le 0$ for all $j\in\mathcal{S}_t$ (opposing or orthogonal outlier), its induced radial weight contribution is zero, whereas Euclidean cluster assignment adds $+1$ to $|C_{\sigma(i)}|$.
\end{proposition}

Next, we connect $\rho_t^2$ directly to nonconvex optimization convergence under preconditioned SGD updates $w_{t+1} = w_t - \eta_t M_t \hat{g}_t$, where $M_t$ satisfies uniform spectral bounds $\mu_M I \preceq M_t \preceq L_M I$ ($0 < \mu_M \le L_M$) and is measurable with respect to the history $\mathcal{F}_t$ prior to sampling $\mathcal{B}_t$. We adopt standard nonconvex smoothness and unbiased candidate-batch assumptions~\citep{ghadimi2013stochastic,bottou2018optimization}.

\begin{assumption}[$L$-smoothness and unbiased batch]
\label{asm:smooth}
The objective $f$ is $L$-smooth and bounded below by $f^\star$. At each step $t$, $\E_t[g_t] = \nabla f(w_t)$ and $\E_t[\|g_t - \nabla f(w_t)\|_{M_t}^2] \le \sigma_M^2 / b$.
\end{assumption}

\begin{theorem}[Preconditioned nonconvex convergence]
\label{thm:nonconvex}
Suppose Assumption~\ref{asm:smooth} holds and $w_{t+1} = w_t - \eta_t M_t \hat{g}_t$ uses deterministic step sizes $0 < \eta_t \le 1/(4 L L_M)$ and metric bounds $\mu_M I \preceq M_t \preceq L_M I$. Define $\Delta_f := f(w_0) - f^\star$ and $e_t := \hat{g}_t - g_t$. Then
\begin{equation}
\begin{aligned}
\frac{\sum_{t=0}^{T-1}\eta_t \E\|\nabla f(w_t)\|_{M_t}^2}{\sum_{t=0}^{T-1}\eta_t}
&\le \frac{2\Delta_f}{\sum_{t=0}^{T-1}\eta_t} \\
&\quad + \frac{2 L L_M \sigma_M^2}{b}\frac{\sum_{t=0}^{T-1}\eta_t^2}{\sum_{t=0}^{T-1}\eta_t} \\
&\quad + \frac{5}{2}\frac{\sum_{t=0}^{T-1}\eta_t \E\|e_t\|_{M_t}^2}{\sum_{t=0}^{T-1}\eta_t}.
\end{aligned}
\label{eq:thm_conv}
\end{equation}
\end{theorem}

\begin{corollary}[Convergence of \method{}]
\label{cor:prism_conv}
Under Theorem~\ref{thm:nonconvex}, substituting Proposition~\ref{prop:radial_bound} into~\eqref{eq:thm_conv} with constant step size $\eta_t = \eta = \sqrt{\Delta_f b / (L L_M \sigma_M^2 T)}$ (for $T$ sufficiently large) yields
\begin{equation}
\begin{aligned}
\frac{1}{T}\sum_{t=0}^{T-1}\E\|\nabla f(w_t)\|_{M_t}^2 &\le 4\sqrt{\frac{\Delta_f L L_M \sigma_M^2}{b T}} + \frac{5}{2}\,\delta_T^2,\\
\delta_T^2 &:= \frac{1}{b T}\sum_{t=0}^{T-1}\E\big[c_t^{\mathrm{cone}}(\mathcal{S}_t)\big].
\end{aligned}
\label{eq:cor_rate}
\end{equation}
\end{corollary}

Setting $\mathcal{S}_t = \mathcal{B}_t$ sends $c_t^{\mathrm{cone}}(\mathcal{S}_t) \to 0$ and recovers the $O(1/\sqrt{bT})$ mini-batch preconditioned SGD rate~\citep{ghadimi2013stochastic}. Under an additional Polyak--{\L}ojasiewicz condition~\citep{karimi2016linear}, Appendix~\ref{app:pl} proves linear contraction to a neighborhood governed by $\sigma_M^2/b$ and $\delta^2$. By contrast, Appendix~\ref{app:nonsubmod} shows via an explicit counterexample that Taylor-approximated validation surrogates used in validation-guided selectors are not submodular in general, leaving their greedy construction without a $(1-1/e)$ certificate.

\begin{figure*}[t]
    \centering
    \includegraphics[width=\textwidth]{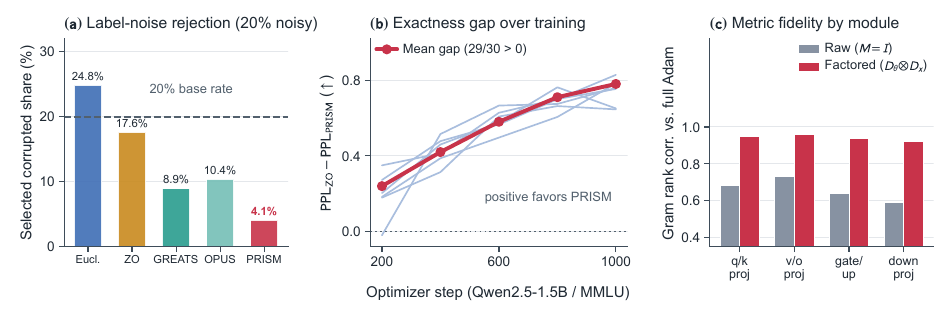}
    \caption{\textbf{Mechanism analysis of \method{}.} (a) Under $20\%$ corrupted responses on LESS/Mistral-7B, Euclidean-FL over-selects corrupted points ($24.8\%$), whereas \method{} suppresses them to $4.1\%$ via zeroed radial projections. (b) Paired test perplexity gap ($\mathrm{PPL}_{\mathrm{ZO}} - \mathrm{PPL}_{\mathrm{PRISM}}$) across 6 subject--seed runs on Qwen2.5-1.5B/MMLU: exact conic geometry outperforms ZeroOrder in $29/30$ nonzero checkpoints. (c) Rank correlation of ghost Gram entries against exact Adam-preconditioned inner products across transformer modules.}
    \label{fig:mechanism}
\end{figure*}

\section{Experiments}
\label{sec:experiments}

We evaluate four questions: (1) Can validation-free conic ghost matching rival or beat validation-directed selectors and full-batch training? (2) Do conic ray similarity, radial weights, factored preconditioning, and exact ghost geometry each contribute gains? (3) Is \method{} computationally practical across scales? (4) When can selection initially lag full-batch training?

\subsection{Experimental Setup}
\label{sec:exp-setup}

\paragraph{Configurations and tasks.}
We evaluate four LoRA fine-tuning setups ($r{=}8, \alpha{=}16$, all linear modules)~\citep{hu2022lora,zou2025flylora,tang2026merge}: (1) \textbf{Qwen2.5-1.5B}~\citep{yang2024qwen25} on \textbf{LESS}~\citep{xia2024less} $\to$ \textbf{MMLU} PPL/Acc~\citep{hendrycks2021measuring} (1{,}000 steps); (2) \textbf{Llama-3.1-8B}~\citep{grattafiori2024llama} on \textbf{Tulu-2}~\citep{ivison2023camels} $\to$ \textbf{BBH} EM~\citep{suzgun2023challenging} (800 steps); (3) \textbf{Mistral-7B}~\citep{jiang2023mistral} on \textbf{LESS} $\to$ \textbf{TyDiQA} F1~\citep{clark2020tydi} (300 steps); and (4) \textbf{Qwen2.5-7B}~\citep{yang2024qwen25} on \textbf{Tulu-2} $\to$ \textbf{GSM8K} Acc~\citep{cobbe2021training} (800 steps). We also test a cold-start setup on \textbf{Qwen3-8B-Base}~\citep{yang2025qwen3} over \textbf{SAMSum}~\citep{gliwa2019samsum} (300 vs.\ 1{,}000 steps). Selectors keep $k{=}8$ of $b{=}16$ candidates ($\rho{=}0.5$); hyperparameters appear in Appendix~\ref{app:setup}.

\paragraph{Baselines.}
We compare against full-batch \textbf{Regular} ($b{=}16$), \textbf{Random} ($k{=}8$), validation/proxy-guided \textbf{GREATS}~\citep{wang2024greats}, \textbf{OPUS}~\citep{wang2026opus}, and \textbf{Filter-then-Weight}~\citep{wang2026filter} (5 held-out target examples), validation-free \textbf{Euclidean-FL} (diameter-shifted Euclidean FL over unpreconditioned ghost gradients with $|C_j|$ weights), and \textbf{ZeroOrder} (CoLM-style sketches~\citep{nguyen2025minibatch} with conic selection). UDS~\citep{zou2025utility}, SLAP~\citep{zou2026slap}, and BLADE~\citep{wang2026blade} appear in Appendix~\ref{app:extra_exp}.

\subsection{Main Results: Validation-Free Selection}
\label{sec:exp-main}

\begin{figure*}[t]
    \centering
    \includegraphics[width=\textwidth]{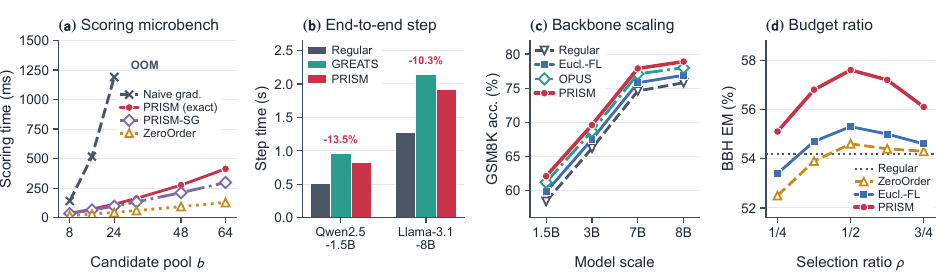}
    \caption{\textbf{Efficiency, scaling, and budget sensitivity.} (a) Scoring + selection time on Qwen2.5-3B (A100): naive gradients OOM past $b{=}24$, while \method{} and \method{}-SG scale to $b{=}64$. (b) Step time ($b{=}16,k{=}8$): \method{} is $13.5\%$ and $10.3\%$ faster than GREATS. (c) GSM8K scaling (1.5B--8B). (d) BBH EM vs.\ selection ratio $\rho{=}k/b$.}
    \label{fig:results}
\end{figure*}

Table~\ref{tab:main} reports downstream results. Without validation or proxy data, \method{} achieves the strongest validation-free score across all four configurations and beats Regular on $4/4$ tasks while updating on half of each batch. On Llama-3.1-8B (BBH) and Qwen2.5-7B (GSM8K), \method{} attains the highest overall accuracy ($57.6\%$ and $77.9\%$), surpassing OPUS ($56.9\%$, $77.1\%$) and GREATS ($56.4\%$, $76.8\%$). On Mistral-7B (TyDiQA), \method{} reaches $0.631$ F1, ahead of OPUS ($0.584$), Filter-then-Weight ($0.597$), and Euclidean-FL ($0.602$), trailing GREATS ($0.654$) by $0.023$. On Qwen2.5-1.5B (MMLU), where GREATS and OPUS observe held-out MMLU examples, \method{} cuts perplexity from $5.84$ (Regular) and $5.61$ (Euclidean-FL) to $5.14$. Complementary accuracy and per-seed breakdowns appear in Appendix~\ref{app:extra_exp}.

\begin{table}[t]
\centering
\footnotesize
\setlength{\tabcolsep}{3.0pt}
\begin{tabular}{l ccc}
\toprule
\textbf{Variant} & \textbf{BBH} & \textbf{TyDiQA} & \textbf{Rel.\ Err.} \\
 & \textbf{EM ($\uparrow$)} & \textbf{F1 ($\uparrow$)} & ($\downarrow$) \\
\midrule
\rowcolor{oursbg}
\textbf{\method{} (full)} & \textbf{57.6} & \textbf{0.631} & \textbf{0.54} \\
\midrule
w/o Radial weights ($|C_j|$) & 56.4 & 0.614 & 0.69 \\
w/o Factored metric ($M{=}I$) & 56.1 & 0.611 & 0.73 \\
w/o Conic sim.\ ($m_t - d_{ij}$) & 55.8 & 0.608 & 0.78 \\
Euclidean-FL (raw, $|C_j|$) & 55.3 & 0.602 & 0.85 \\
ZeroOrder (ZO geometry) & 54.6 & 0.579 & 0.94 \\
\midrule
\method{}-SG ($\varepsilon{=}0.1$) & 57.3 & 0.627 & 0.57 \\
\bottomrule
\end{tabular}
\caption{\textbf{Component ablations} on Llama-3.1-8B (BBH) and Mistral-7B (TyDiQA), with relative gradient error $\|\hat{g}_t - g_t\|_M / \|g_t\|_M$.}
\label{tab:ablation}
\end{table}

\subsection{Why Conic Geometry and Preconditioning Matter}
\label{sec:exp-ablation}

Table~\ref{tab:ablation} isolates each component. Replacing radial weights~\eqref{eq:radial_weights} with integer counts $|C_j|$ drops BBH by $1.2$ points ($57.6 \to 56.4$) and raises relative gradient error from $0.54$ to $0.69$. Removing factored preconditioning ($M{=}I$) drops BBH by $1.5$ points ($56.1$) due to MLP scale anisotropy. Reverting conic similarity to diameter-shifted Euclidean similarity ($m_t - d_{ij}$) loses $1.8$ points ($55.8$) via outlier diameter inflation (Figure~\ref{fig:motivation}a). Combining all three regressions into Euclidean-FL drops $2.3$ points ($55.3$), and ZeroOrder drops $3.0$ points ($54.6$). Finally, \method{}-SG ($\varepsilon{=}0.1$) stays within $0.3$ points ($57.3$).

\subsection{Mechanism Analysis}
\label{sec:exp-mechanism}

\paragraph{Noise rejection via radial projection.}
Corrupting $20\%$ of LESS (Mistral-7B) targets, Figure~\ref{fig:mechanism}(a) shows Euclidean-FL picks corrupted items at $24.8\%$ (above the $20\%$ base rate) as outlier norms inflate $m_t$, whereas \method{} cuts them to $4.1\%$---under half of GREATS ($8.9\%$) and OPUS ($10.4\%$)---via~\eqref{eq:radial_weights}.

\paragraph{Exact ghost geometry vs.\ zeroth-order.}
Figure~\ref{fig:mechanism}(b) shows \method{} beats ZeroOrder in $29/30$ nonzero checkpoints on Qwen2.5-1.5B (MMLU) and $10/10$ on Mistral-7B TyDiQA (Appendix~\ref{app:extra_exp}), confirming Corollary~\ref{cor:prism_conv}.

\paragraph{Metric fidelity.}
On Llama-3.1-8B (Figure~\ref{fig:mechanism}c), raw ghost inner products ($M{=}I$) correlate with diagonal AdamW at $0.59$--$0.73$, whereas $D_\theta \otimes D_x$ reaches $0.92$--$0.96$.

\subsection{Efficiency, Scaling, and Cold-Start Boundary}
\label{sec:exp-efficiency}

\paragraph{Computational efficiency.}
On Qwen2.5-3B (A100, Figure~\ref{fig:results}a), naive gradients take $1{,}190$\,ms at $b{=}24$ and OOM by $b{=}32$; \method{} runs in $115$\,ms ($10.3\times$ faster) and scales to $b{=}64$ ($414$\,ms; $298$\,ms with \method{}-SG). End-to-end (Figure~\ref{fig:results}b), \method{} runs in $0.83$\,s vs.\ $0.52$\,s Regular on Qwen2.5-1.5B and $1.92$\,s vs.\ $1.28$\,s on Llama-3.1-8B, beating GREATS by $13.5\%$ and $10.3\%$.

\paragraph{Scaling and selection ratio.}
\method{}'s gains hold from 1.5B to 8B on GSM8K (Figure~\ref{fig:results}c), and beat Regular across $\rho \in [0.25, 0.75]$ on BBH, peaking at $\rho{=}0.50$ (Figure~\ref{fig:results}d).

\paragraph{Cold-start boundary case.}
Fine-tuning raw \textbf{Qwen3-8B-Base} on \textbf{SAMSum} with an unseen chat template for $300$ steps, Regular ($0.214$ ROUGE-1) leads OPUS ($0.189$), GREATS ($0.172$), \method{} ($0.148$), Euclidean-FL ($0.131$), and ZeroOrder ($0.084$), as early format acquisition favors raw volume. By $1{,}000$ steps (Figure~\ref{fig:app_horizons}c), \method{} reaches $0.376$, ahead of GREATS ($0.369$), OPUS ($0.367$), and Regular ($0.341$).

\section{Conclusion}
\label{sec:conclusion}

We introduced \method{}, a validation-free online batch selector using conic facility location over factored preconditioned ghost inner products. Replacing diameter-shifted Euclidean distances with nonnegative ray similarities avoids outlier inflation while preserving the $(1-1/e)$ submodular guarantee, and closed-form radial weights damp noisy examples. Nonconvex and PL bounds link optimization error directly to conic cost. Across four LLM fine-tuning tasks (1.5B--8B), \method{} outperforms full-batch training, Euclidean-FL, and zeroth-order selection, rivals validation-guided baselines without held-out data, and rejects label noise.

\section*{Limitations}

\method{} matches the geometry of the current candidate batch, so when the target evaluation distribution differs sharply from the training pool and a reliable target validation set is available, validation-guided selectors can exploit information absent from the batch. Our convergence theorem assumes smooth objectives, fixed-horizon or diminishing step sizes under preconditioned SGD, and bounded metric eigenvalues, whereas our experiments use AdamW; extending the analysis directly to stateful adaptive momentum dynamics remains open. In addition, although factored ghost inner products avoid storing $p$-dimensional per-example gradients, exact $b\times b$ Gram computation still scales quadratically in the candidate pool size $b$ and adds memory proportional to batch activations, making stochastic greedy and moderate pool sizes ($b \le 64$) preferable on single-GPU setups. Finally, in extreme cold-start regimes where a base model must rapidly acquire a novel output template within a few hundred steps, warming up on full batches before enabling selection is advantageous.

\section*{Ethics Statement}

\method{} alters only which training examples enter each parameter update and uses public instruction-tuning and reasoning benchmarks under their original licenses. No human subjects or private user data are involved. Like other online batch selectors, down-weighting gradients that diverge from the batch majority is beneficial for filtering mislabeled or noisy examples, but could also under-weight rare yet legitimate minority phenomena (such as underrepresented languages or dialects in multi-source pools) if deployed without per-source stratification; practitioners applying \method{} to sensitive or heterogeneous corpora should monitor per-source retention rates.

\bibliography{custom}

@article{loshchilov2015online,
  title={Online Batch Selection for Faster Training of Neural Networks},
  author={Loshchilov, Ilya and Hutter, Frank},
  journal={arXiv preprint arXiv:1511.06343},
  year={2015},
  url={https://arxiv.org/abs/1511.06343}
}

@inproceedings{katharopoulos2018not,
  title={Not All Samples Are Created Equal: Deep Learning with Importance Sampling},
  author={Katharopoulos, Angelos and Fleuret, Fran{\c{c}}ois},
  booktitle={Proceedings of the 35th International Conference on Machine Learning},
  pages={2525--2534},
  year={2018},
  organization={PMLR},
  url={https://arxiv.org/abs/1803.00942}
}

@inproceedings{mirzasoleiman2015lazier,
  title={Lazier Than Lazy Greedy},
  author={Mirzasoleiman, Baharan and Badanidiyuru, Ashwinkumar and Karbasi, Amin and Vondrak, Jan and Krause, Andreas},
  booktitle={Proceedings of the AAAI Conference on Artificial Intelligence},
  volume={29},
  year={2015},
  url={https://arxiv.org/abs/1409.7938}
}

@inproceedings{mirzasoleiman2020coresets,
  title={Coresets for Data-efficient Training of Machine Learning Models},
  author={Mirzasoleiman, Baharan and Bilmes, Jeff and Leskovec, Jure},
  booktitle={Proceedings of the 37th International Conference on Machine Learning},
  pages={6950--6960},
  year={2020},
  organization={PMLR},
  url={https://arxiv.org/abs/1906.01827}
}

@inproceedings{killamsetty2021gradmatch,
  title={{GRAD-MATCH}: Gradient Matching based Data Subset Selection for Efficient Deep Model Training},
  author={Killamsetty, Krishnateja and Sivasubramanian, Durga and Ramakrishnan, Ganesh and De, Abir and Iyer, Rishabh},
  booktitle={Proceedings of the 38th International Conference on Machine Learning},
  pages={5464--5474},
  year={2021},
  organization={PMLR},
  url={https://arxiv.org/abs/2103.00123}
}

@inproceedings{hu2022lora,
  title={{LoRA}: Low-Rank Adaptation of Large Language Models},
  author={Hu, Edward J. and Shen, Yelong and Wallis, Phillip and Allen-Zhu, Zeyuan and Li, Yuanzhi and Wang, Shean and Wang, Lu and Chen, Weizhu},
  booktitle={International Conference on Learning Representations},
  year={2022},
  url={https://arxiv.org/abs/2106.09685}
}

@inproceedings{mindermann2022prioritized,
  title={Prioritized Training on Points that are Learnable, Worth Learning, and Not Yet Learnt},
  author={Mindermann, S{\"o}ren and Brauner, Jan and Razzak, Muhammed and Sharma, Mrinank and Kirsch, Andreas and Xu, Winnie and H{\"o}ltgen, Benedikt and Gomez, Aidan N. and Morisot, Adrien and Farquhar, Sebastian and Gal, Yarin},
  booktitle={Proceedings of the 39th International Conference on Machine Learning},
  pages={15630--15649},
  year={2022},
  organization={PMLR},
  url={https://arxiv.org/abs/2206.07137}
}

@inproceedings{li2022large,
  title={Large Language Models Can Be Strong Differentially Private Learners},
  author={Li, Xuechen and Tram{\`e}r, Florian and Liang, Percy and Hashimoto, Tatsunori},
  booktitle={International Conference on Learning Representations},
  year={2022},
  url={https://arxiv.org/abs/2110.05679}
}

@inproceedings{zhang2020why,
  title={Why are Adaptive Methods Good for Attention Models?},
  author={Zhang, Jingzhao and Karimireddy, Sai Praneeth and Veit, Andreas and Kim, Seungyeon and Reddi, Sashank J and Kumar, Sanjiv and Sra, Suvrit},
  booktitle={Advances in Neural Information Processing Systems},
  volume={33},
  pages={15383--15393},
  year={2020},
  url={https://arxiv.org/abs/1912.03194}
}

@inproceedings{hendrycks2021measuring,
  title={Measuring Massive Multitask Language Understanding},
  author={Hendrycks, Dan and Burns, Collin and Basart, Steven and Zou, Andy and Mazeika, Mantas and Song, Dawn and Steinhardt, Jacob},
  booktitle={International Conference on Learning Representations},
  year={2021},
  url={https://arxiv.org/abs/2009.03300}
}

@article{clark2020tydi,
  title={{TyDi} {QA}: A Benchmark for Information-Seeking Question Answering in Typologically Diverse Languages},
  author={Clark, Jonathan H. and Choi, Eunsol and Collins, Michael and Garrette, Dan and Kwiatkowski, Tom and Nikolaev, Vitaly and Palomaki, Jennimaria},
  journal={Transactions of the Association for Computational Linguistics},
  volume={8},
  pages={454--470},
  year={2020},
  url={https://arxiv.org/abs/2003.05002}
}

@inproceedings{suzgun2023challenging,
  title={Challenging {BIG-Bench} Tasks and Whether Chain-of-Thought Can Solve Them},
  author={Suzgun, Mirac and Scales, Nathan and Sch{\"a}rli, Nathanael and Gehrmann, Sebastian and Tay, Yi and Chung, Hyung Won and Chowdhery, Aakanksha and Le, Quoc V. and Chi, Ed H. and Zhou, Denny and Wei, Jason},
  booktitle={Findings of the Association for Computational Linguistics: ACL 2023},
  pages={13003--13051},
  year={2023},
  url={https://arxiv.org/abs/2210.09261}
}

@article{cobbe2021training,
  title={Training Verifiers to Solve Math Word Problems},
  author={Cobbe, Karl and Kosaraju, Vineet and Bavarian, Mohammad and Chen, Mark and Jun, Heewoo and Kaiser, Lukasz and Plappert, Matthias and Tworek, Jerry and Hilton, Jacob and Nakano, Reiichiro and Hesse, Christopher and Schulman, John},
  journal={arXiv preprint arXiv:2110.14168},
  year={2021},
  url={https://arxiv.org/abs/2110.14168}
}

@inproceedings{gliwa2019samsum,
  title={{SAMSum} Corpus: A Human-annotated Dialogue Dataset for Abstractive Summarization},
  author={Gliwa, Bogdan and Mochol, Iwona and Biesek, Maciej and Wawer, Aleksander},
  booktitle={Proceedings of the 2nd Workshop on New Frontiers in Summarization},
  pages={70--79},
  year={2019},
  url={https://arxiv.org/abs/1911.12237}
}

@article{jiang2023mistral,
  title={{Mistral} {7B}},
  author={Jiang, Albert Q. and Sablayrolles, Alexandre and Mensch, Arthur and Bamford, Chris and Chaplot, Devendra Singh and Casas, Diego de las and Bressand, Florian and Lengyel, Gianna and Lample, Guillaume and Saulnier, Lucile and Lavaud, L{\'e}lio Renard and Lachaux, Marie-Anne and Stock, Pierre and Scao, Teven Le and Lavril, Thibaut and Wang, Thomas and Lacroix, Timoth{\'e}e and Sayed, William El},
  journal={arXiv preprint arXiv:2310.06825},
  year={2023},
  url={https://arxiv.org/abs/2310.06825}
}

@article{ivison2023camels,
  title={Camels in a Changing Climate: Enhancing {LM} Adaptation with {Tulu} 2},
  author={Ivison, Hamish and Wang, Yizhong and Pyatkin, Valentina and Lambert, Nathan and Peters, Matthew and Dasigi, Pradeep and Jang, Joel and Wadden, David and Smith, Noah A. and Beltagy, Iz and Hajishirzi, Hannaneh},
  journal={arXiv preprint arXiv:2311.10702},
  year={2023},
  url={https://arxiv.org/abs/2311.10702}
}

@inproceedings{xia2024less,
  title={LESS: Selecting Influential Data for Targeted Instruction Tuning},
  author={Xia, Mengzhou and Malladi, Sadhika and Gururangan, Suchin and Arora, Sanjeev and Chen, Danqi},
  booktitle={Proceedings of the 41st International Conference on Machine Learning},
  pages={54104--54132},
  year={2024},
  organization={PMLR},
  url={https://arxiv.org/abs/2402.04333}
}

@inproceedings{evans2024data,
  title={Data curation via joint example selection further accelerates multimodal learning},
  author={Evans, Talfan and Parthasarathy, Nikhil and Merzic, Hamza and Henaff, Olivier J.},
  booktitle={Advances in Neural Information Processing Systems},
  volume={37},
  year={2024},
  url={https://arxiv.org/abs/2406.17711}
}

@article{grattafiori2024llama,
  title={The {Llama} 3 Herd of Models},
  author={Grattafiori, Aaron and Dubey, Abhimanyu and Jauhri, Abhinav and Pandey, Abhinav and Kadian, Abhishek and Al-Dahle, Ahmad and Letman, Aiesha and Mathur, Akhil and Schelten, Alan and Vaughan, Alex and Yang, Amy and Fan, Angela and Goyal, Anirudh and Hartshorn, Anthony and Yang, Aobo and Mitra, Archi and Sravankumar, Archie and Korenev, Artem and Hinsvark, Arthur and Rao, Arun and Zhang, Aston and Rodriguez, Aurelien and Gregerson, Austen and Spataru, Ava and Roziere, Baptiste and Biron, Bethany and Tang, Binh and Chern, Bobbie and Caucheteux, Charlotte and Nayak, Chaya and Bi, Chloe and Marra, Chris and McConnell, Chris and Keller, Christian and Touret, Christophe and Wu, Chunyang and Wong, Corinne and Ferrer, Cristian Canton and Nikolaidis, Cyrus and Allonsius, Damien and Song, Daniel and Pintz, Danielle and Livshits, Danny and Wyatt, Danny and Esiobu, David and Choudhary, Dhruv and Mahajan, Dhruv and Garcia-Olano, Diego and Perino, Diego and Hupkes, Dieuwke and Lakomkin, Egor and AlBadawy, Ehab and Lobanova, Elina and Dinan, Emily and Smith, Eric Michael and Radenovic, Filip and Guzm{\'a}n, Francisco and Zhang, Frank and Synnaeve, Gabriel and Lee, Gabrielle and Anderson, Georgia Lewis and Thattai, Govind and Nail, Graeme and Mialon, Gregoire and Pang, Guan and Cucurell, Guillem and Nguyen, Hailey and Korevaar, Hannah and Xu, Hu and Touvron, Hugo and Zarov, Iliyan and Ibarra, Imanol Arrieta and Kloumann, Isabel and Misra, Ishan and Evtimov, Ivan and Zhang, Jack and Copet, Jade and Lee, Jaewon and Geffert, Jan and Vranes, Jana and Park, Jason and Mahadeokar, Jay and Shah, Jeet and Linde, Jelmer van der and Billock, Jennifer and Hong, Jenny and Lee, Jenya and Fu, Jeremy and Chi, Jianfeng and Huang, Jianyu and Liu, Jiawen and Wang, Jie and Yu, Jiecao and Bitton, Joanna and Spisak, Joe and Park, Jongsoo and Rocca, Joseph and Johnstun, Joshua and Saxe, Joshua and Jia, Junteng and Alwala, Kalyan Vasuden and Prasad, Karthik and Upasani, Kartikeya and Plawiak, Kate and Li, Ke and Heafield, Kenneth and Stone, Kevin and El-Arini, Khalid and Iyer, Krithika and Malik, Kshitiz and Chiu, Kuenley and Bhalla, Kunal and Lakhotia, Kushal and Rantala-Yeary, Lauren and Maaten, Laurens van der and Chen, Lawrence and Tan, Liang and Jenkins, Liz and Martin, Louis and Madaan, Lovish and Malo, Lubo and Blecher, Lukas and Landzaat, Lukas and Oliveira, Luke de and Muzzi, Madeline and Pasupuleti, Mahesh and Singh, Mannat and Paluri, Manohar and Kardas, Marcin and Tsimpoukelli, Maria and Oldham, Mathew and Rita, Mathieu and Pavlova, Maya and Kambadur, Melanie and Lewis, Mike and Si, Min and Singh, Mitesh Kumar and Hassan, Mona and Goyal, Naman and Torabi, Narjes and Bashlykov, Nikolay and Bogoychev, Nikolay and Chatterji, Niladri and Zhang, Ning and Duchenne, Olivier and {\c{C}}elebi, Onur and Alrassy, Patrick and Zhang, Pengchuan and Li, Pengwei and Vasic, Petar and Weng, Peter and Bhargava, Prajjwal and Dubal, Pratik and Krishnan, Praveen and Koura, Punit Singh and Xu, Puxin and He, Qing and Dong, Qingxiao and Srinivasan, Ragavan and Ganapathy, Raj and Calderer, Ramon and Cabral, Ricardo Silveira and Stojnic, Robert and Raileanu, Roberta and Maheswari, Rohan and Girdhar, Rohit and Patel, Rohit and Sauvestre, Romain and Polidoro, Ronnie and Sumbaly, Roshan and Taylor, Ross and Silva, Ruan and Hou, Rui and Wang, Rui and Hosseini, Saghar and Chennabasappa, Sahana and Singh, Sanjay and Bell, Sean and Kim, Seohyun Sonia and Edunov, Sergey and Nie, Shaoliang and Narang, Sharan and Raparthy, Sharath and Shen, Sheng and Wan, Shengye and Bhosale, Shruti and Zhang, Shun and Vandenhende, Simon and Batra, Soumya and Whitman, Spencer and Sootla, Sten and Collot, Stephane and Gururangan, Suchin and Borodinsky, Sydney and Herman, Tamar and Fowler, Tara and Sheasha, Tarek and Georgiou, Thomas and Scialom, Thomas and Speckbacher, Tobias and Mihaylov, Todor and Xiao, Tong and Karn, Ujjwal and Goswami, Vedanuj and Gupta, Vibhor and Ramanathan, Vignesh and Kerkez, Viktor and Gonguet, Vincent and Do, Virginie and Vogeti, Vish and Albiero, V{\'i}tor and Petrovic, Vladan and Chu, Weiwei and Xiong, Wenhan and Fu, Wenyin and Meers, Whitney and Martinet, Xavier and Wang, Xiaodong and Wang, Xiaofang and Tan, Xiaoqing Ellen and Xia, Xide and Xie, Xinfeng and Jia, Xuchao and Wang, Xuewei and Goldschlag, Yaelle and Gaur, Yashesh and Babaei, Yasmine and Wen, Yi and Song, Yiwen and Zhang, Yuchen and Li, Yue and Mao, Yuning and Coudert, Zacharie Delpierre and Yan, Zheng and Chen, Zhengxing and Papakipos, Zoe and Singh, Aaditya and Srivastava, Aayushi and Jain, Abha and Kelsey, Adam and Shajnfeld, Adam and Gangidi, Adithya and Victoria, Adolfo and Goldstand, Ahuva and Menon, Ajay and Sharma, Ajay and Boesenberg, Alex and Baevski, Alexei and Feinstein, Allie and Kallet, Amanda and Sangani, Amit and Teo, Amos and Yunus, Anam and Lupu, Andrei and Alvarado, Andres and Caples, Andrew and Gu, Andrew and Ho, Andrew and Poulton, Andrew and Ryan, Andrew and Ramchandani, Ankit and Dong, Annie and Franco, Annie and Goyal, Anuj and Saraf, Aparajita and Chowdhury, Arkabandhu and Gabriel, Ashley and Bharambe, Ashwin and Eisenman, Assaf and Yazdan, Azadeh and James, Beau and Maurer, Ben and Leonhardi, Benjamin and Huang, Bernie and Loyd, Beth and Paola, Beto De and Paranjape, Bhargavi and Liu, Bing and Wu, Bo and Ni, Boyu and Hancock, Braden and Wasti, Bram and Spence, Brandon and Stojkovic, Brani and Gamido, Brian and Montalvo, Britt and Parker, Carl and Burton, Carly and Mejia, Catalina and Liu, Ce and Wang, Changhan and Kim, Changkyu and Zhou, Chao and Hu, Chester and Chu, Ching-Hsiang and Cai, Chris and Tindal, Chris and Feichtenhofer, Christoph and Gao, Cynthia and Civin, Damon and Beaty, Dana and Kreymer, Daniel and Li, Daniel and Adkins, David and Xu, David and Testuggine, Davide and David, Delia and Parikh, Devi and Liskovich, Diana and Foss, Didem and Wang, Dingkang and Le, Duc and Holland, Dustin and Dowling, Edward and Jamil, Eissa and Montgomery, Elaine and Presani, Eleonora and Hahn, Emily and Wood, Emily and Le, Eric-Tuan and Brinkman, Erik and Arcaute, Esteban and Dunbar, Evan and Smothers, Evan and Sun, Fei and Kreuk, Felix and Tian, Feng and Kokkinos, Filippos and Ozgenel, Firat and Caggioni, Francesco and Kanayet, Frank and Seide, Frank and Florez, Gabriela Medina and Schwarz, Gabriella and Badeer, Gada and Swee, Georgia and Halpern, Gil and Herman, Grant and Sizov, Grigory and {Guangyi} and {Zhang} and Lakshminarayanan, Guna and Inan, Hakan and Shojanazeri, Hamid and Zou, Han and Wang, Hannah and Zha, Hanwen and Habeeb, Haroun and Rudolph, Harrison and Suk, Helen and Aspegren, Henry and Goldman, Hunter and Zhan, Hongyuan and Damlaj, Ibrahim and Molybog, Igor and Tufanov, Igor and Leontiadis, Ilias and Veliche, Irina-Elena and Gat, Itai and Weissman, Jake and Geboski, James and Kohli, James and Lam, Janice and Asher, Japhet and Gaya, Jean-Baptiste and Marcus, Jeff and Tang, Jeff and Chan, Jennifer and Zhen, Jenny and Reizenstein, Jeremy and Teboul, Jeremy and Zhong, Jessica and Jin, Jian and Yang, Jingyi and Cummings, Joe and Carvill, Jon and Shepard, Jon and McPhie, Jonathan and Torres, Jonathan and Ginsburg, Josh and Wang, Junjie and Wu, Kai and U, Kam Hou and Saxena, Karan and Khandelwal, Kartikay and Zand, Katayoun and Matosich, Kathy and Veeraraghavan, Kaushik and Michelena, Kelly and Li, Keqian and Jagadeesh, Kiran and Huang, Kun and Chawla, Kunal and Huang, Kyle and Chen, Lailin and Garg, Lakshya and A, Lavender and Silva, Leandro and Bell, Lee and Zhang, Lei and Guo, Liangpeng and Yu, Licheng and Moshkovich, Liron and Wehrstedt, Luca and Khabsa, Madian and Avalani, Manav and Bhatt, Manish and Mankus, Martynas and Hasson, Matan and Lennie, Matthew and Reso, Matthias and Groshev, Maxim and Naumov, Maxim and Lathi, Maya and Keneally, Meghan and Liu, Miao and Seltzer, Michael L. and Valko, Michal and Restrepo, Michelle and Patel, Mihir and Vyatskov, Mik and Samvelyan, Mikayel and Clark, Mike and Macey, Mike and Wang, Mike and Hermoso, Miquel Jubert and Metanat, Mo and Rastegari, Mohammad and Bansal, Munish and Santhanam, Nandhini and Parks, Natascha and White, Natasha and Bawa, Navyata and Singhal, Nayan and Egebo, Nick and Usunier, Nicolas and Mehta, Nikhil and Laptev, Nikolay Pavlovich and Dong, Ning and Cheng, Norman and Chernoguz, Oleg and Hart, Olivia and Salpekar, Omkar and Kalinli, Ozlem and Kent, Parkin and Parekh, Parth and Saab, Paul and Balaji, Pavan and Rittner, Pedro and Bontrager, Philip and Roux, Pierre and Dollar, Piotr and Zvyagina, Polina and Ratanchandani, Prashant and Yuvraj, Pritish and Liang, Qian and Alao, Rachad and Rodriguez, Rachel and Ayub, Rafi and Murthy, Raghotham and Nayani, Raghu and Mitra, Rahul and Parthasarathy, Rangaprabhu and Li, Raymond and Hogan, Rebekkah and Battey, Robin and Wang, Rocky and Howes, Russ and Rinott, Ruty and Mehta, Sachin and Siby, Sachin and Bondu, Sai Jayesh and Datta, Samyak and Chugh, Sara and Hunt, Sara and Dhillon, Sargun and Sidorov, Sasha and Pan, Satadru and Mahajan, Saurabh and Verma, Saurabh and Yamamoto, Seiji and Ramaswamy, Sharadh and Lindsay, Shaun and Lindsay, Shaun and Feng, Sheng and Lin, Shenghao and Zha, Shengxin Cindy and Patil, Shishir and Shankar, Shiva and Zhang, Shuqiang and Zhang, Shuqiang and Wang, Sinong and Agarwal, Sneha and Sajuyigbe, Soji and Chintala, Soumith and Max, Stephanie and Chen, Stephen and Kehoe, Steve and Satterfield, Steve and Govindaprasad, Sudarshan and Gupta, Sumit and Deng, Summer and Cho, Sungmin and Virk, Sunny and Subramanian, Suraj and Choudhury, Sy and Goldman, Sydney and Remez, Tal and Glaser, Tamar and Best, Tamara and Koehler, Thilo and Robinson, Thomas and Li, Tianhe and Zhang, Tianjun and Matthews, Tim and Chou, Timothy and Shaked, Tzook and Vontimitta, Varun and Ajayi, Victoria and Montanez, Victoria and Mohan, Vijai and Kumar, Vinay Satish and Mangla, Vishal and Ionescu, Vlad and Poenaru, Vlad and Mihailescu, Vlad Tiberiu and Ivanov, Vladimir and Li, Wei and Wang, Wenchen and Jiang, Wenwen and Bouaziz, Wes and Constable, Will and Tang, Xiaocheng and Wu, Xiaojian and Wang, Xiaolan and Wu, Xilun and Gao, Xinbo and Kleinman, Yaniv and Chen, Yanjun and Hu, Ye and Jia, Ye and Qi, Ye and Li, Yenda and Zhang, Yilin and Zhang, Ying and Adi, Yossi and Nam, Youngjin and {Yu} and {Wang} and Zhao, Yu and Hao, Yuchen and Qian, Yundi and Li, Yunlu and He, Yuzi and Rait, Zach and DeVito, Zachary and Rosnbrick, Zef and Wen, Zhaoduo and Yang, Zhenyu and Zhao, Zhiwei and Ma, Zhiyu},
  journal={arXiv preprint arXiv:2407.21783},
  year={2024},
  url={https://arxiv.org/abs/2407.21783}
}

@article{yang2024qwen25,
  title={{Qwen2.5} Technical Report},
  author={{Qwen} and Yang, An and Yang, Baosong and Zhang, Beichen and Hui, Binyuan and Zheng, Bo and Yu, Bowen and Li, Chengyuan and Liu, Dayiheng and Huang, Fei and Wei, Haoran and Lin, Huan and Yang, Jian and Tu, Jianhong and Zhang, Jianwei and Yang, Jianxin and Yang, Jiaxi and Zhou, Jingren and Lin, Junyang and Dang, Kai and Lu, Keming and Bao, Keqin and Yang, Kexin and Yu, Le and Li, Mei and Xue, Mingfeng and Zhang, Pei and Zhu, Qin and Men, Rui and Lin, Runji and Li, Tianhao and Tang, Tianyi and Xia, Tingyu and Ren, Xingzhang and Ren, Xuancheng and Fan, Yang and Su, Yang and Zhang, Yichang and Wan, Yu and Liu, Yuqiong and Cui, Zeyu and Zhang, Zhenru and Qiu, Zihan},
  journal={arXiv preprint arXiv:2412.15115},
  year={2024},
  url={https://arxiv.org/abs/2412.15115}
}

@article{albalak2024survey,
  title={A Survey on Data Selection for Language Models},
  author={Albalak, Alon and Elazar, Yanai and Xie, Sang Michael and Longpre, Shayne and Lambert, Nathan and Wang, Xinyi and Muennighoff, Niklas and Hou, Bairu and Pan, Liangming and Jeong, Haewon and Raffel, Colin and Chang, Shiyu and Hashimoto, Tatsunori and Wang, William Yang},
  journal={Transactions on Machine Learning Research},
  year={2024},
  url={https://arxiv.org/abs/2402.16827}
}

@inproceedings{nguyen2025minibatch,
  title={Mini-batch Coresets for Memory-efficient Language Model Training on Data Mixtures},
  author={Nguyen, Dang and Yang, Wenhan and Anand, Rathul and Yang, Yu and Mirzasoleiman, Baharan},
  booktitle={The Thirteenth International Conference on Learning Representations},
  year={2025},
  url={https://arxiv.org/abs/2407.19580}
}

@inproceedings{wang2025data,
  title={Data {Shapley} in One Training Run},
  author={Wang, Jiachen T. and Mittal, Prateek and Song, Dawn and Jia, Ruoxi},
  booktitle={The Thirteenth International Conference on Learning Representations},
  year={2025},
  url={https://arxiv.org/abs/2406.11011}
}

@inproceedings{zhang2025tagcos,
  title={{TAGCOS}: Task-agnostic Gradient Clustered Coreset Selection for Instruction Tuning Data},
  author={Zhang, Jipeng and Qin, Yaxuan and Pi, Renjie and Zhang, Weizhong and Pan, Rui and Zhang, Tong},
  booktitle={Findings of the Association for Computational Linguistics: NAACL 2025},
  year={2025},
  url={https://arxiv.org/abs/2407.15235}
}

@inproceedings{tan2025infomax,
  title={Data Pruning by Information Maximization},
  author={Tan, Haoru and Wu, Sitong and Huang, Wei and Zhao, Shizhen and Qi, Xiaojuan},
  booktitle={The Thirteenth International Conference on Learning Representations},
  year={2025},
  url={https://arxiv.org/abs/2506.01701}
}

@article{zou2025utility,
  title={Utility-Diversity Aware Online Batch Selection for {LLM} Supervised Fine-tuning},
  author={Zou, Heming and Mao, Yixiu and Qu, Yun and Wang, Qi and Ji, Xiangyang},
  journal={arXiv preprint arXiv:2510.16882},
  year={2025},
  url={https://arxiv.org/abs/2510.16882}
}

@article{lin2025lead,
  title={{LEAD}: Iterative Data Selection for Efficient {LLM} Instruction Tuning},
  author={Lin, Xiaotian and Qi, Yanlin and Zhu, Yizhang and Palpanas, Themis and Chai, Chengliang and Tang, Nan and Luo, Yuyu},
  journal={arXiv preprint arXiv:2505.07437},
  year={2025},
  url={https://arxiv.org/abs/2505.07437}
}

@article{hu2025donod,
  title={{DONOD}: Efficient and Generalizable Instruction Fine-Tuning for {LLMs} via Model-Intrinsic Dataset Pruning},
  author={Hu, Jucheng and Yang, Surong and Wu, Lijun and Zhou, Dongzhan},
  journal={arXiv preprint arXiv:2504.14810},
  year={2025},
  url={https://arxiv.org/abs/2504.14810}
}

@article{nikdan2025efficient,
  title={Efficient Data Selection at Scale via Influence Distillation},
  author={Nikdan, Mahdi and Cohen-Addad, Vincent and Alistarh, Dan and Mirrokni, Vahab},
  journal={arXiv preprint arXiv:2505.19051},
  year={2025},
  url={https://arxiv.org/abs/2505.19051}
}

@article{yang2025qwen3,
  title={{Qwen3} Technical Report},
  author={Yang, An and Li, Anfeng and Yang, Baosong and Zhang, Beichen and Hui, Binyuan and Zheng, Bo and Yu, Bowen and Gao, Chang and Huang, Chengen and Lv, Chenxu and Zheng, Chujie and Liu, Dayiheng and Zhou, Fan and Huang, Fei and Hu, Feng and Ge, Hao and Wei, Haoran and Lin, Huan and Tang, Jialong and Yang, Jian and Tu, Jianhong and Zhang, Jianwei and Yang, Jianxin and Yang, Jiaxi and Zhou, Jing and Zhou, Jingren and Lin, Junyang and Dang, Kai and Bao, Keqin and Yang, Kexin and Yu, Le and Deng, Lianghao and Li, Mei and Xue, Mingfeng and Li, Mingze and Zhang, Pei and Wang, Peng and Zhu, Qin and Men, Rui and Gao, Ruize and Liu, Shixuan and Luo, Shuang and Li, Tianhao and Tang, Tianyi and Yin, Wenbiao and Ren, Xingzhang and Wang, Xinyu and Zhang, Xinyu and Ren, Xuancheng and Fan, Yang and Su, Yang and Zhang, Yichang and Zhang, Yinger and Wan, Yu and Liu, Yuqiong and Wang, Zekun and Cui, Zeyu and Zhang, Zhenru and Zhou, Zhipeng and Qiu, Zihan},
  journal={arXiv preprint arXiv:2505.09388},
  year={2025},
  url={https://arxiv.org/abs/2505.09388}
}

@article{wang2026opus,
  title={{OPUS}: Towards Efficient and Principled Data Selection in Large Language Model Pre-training in Every Iteration},
  author={Wang, Shaobo and Ouyang, Xuan and Xu, Tianyi and Hu, Yuzheng and Liu, Jialin and Chen, Guo and Zhang, Tianyu and Zheng, Junhao and Yang, Kexin and Ren, Xingzhang and Liu, Dayiheng and Zhang, Linfeng},
  journal={arXiv preprint arXiv:2602.05400},
  year={2026},
  url={https://arxiv.org/abs/2602.05400}
}

@article{wang2026filter,
  title={Filter-then-Weight: Online Data Selection and Reweighting for {LLM} Fine-Tuning},
  author={Wang, Fangxin and Baghershahi, Peyman and He, Langzhou and Zou, Henry Peng and Medya, Sourav and Yu, Philip S.},
  journal={arXiv preprint arXiv:2604.00001},
  year={2026},
  url={https://arxiv.org/abs/2604.00001}
}

@article{wang2026blade,
  title={{BLADE}: Scalable Bi-level Adaptive Data Selection for {LLM} Training},
  author={Wang, Jiaxing and Xiang, Deping and Xu, Jin and Liu, Zirui and Zhang, Zicheng and Gong, Guoqiang and Fang, Jun and Liu, Chao and Liu, Pengzhang and Liu, Tongxuan and Zhang, Ke and Jiang, Qixia},
  journal={arXiv preprint arXiv:2606.18650},
  year={2026},
  url={https://arxiv.org/abs/2606.18650}
}

@article{wang2026gaia,
  title={Online Data Selection for Instruction Tuning via {Gaussian} Processes},
  author={Wang, Jun and Nguyen, Quoc Phong and Monteil, Julien and Nguyen, Vu},
  journal={arXiv preprint arXiv:2606.30077},
  year={2026},
  url={https://arxiv.org/abs/2606.30077}
}

@article{tang2026grace,
  title={{GRACE}: A Dynamic Coreset Selection Framework for Large Language Model Optimization},
  author={Tang, Tianhao and Li, Haoyang and Chen, Lei},
  journal={arXiv preprint arXiv:2604.11810},
  year={2026},
  url={https://arxiv.org/abs/2604.11810}
}

@article{zou2026slap,
  title={{SLAP}: Stratified Loss-based Pruning for On-Policy Data-Efficient Instruction Tuning},
  author={Zou, Run and Ding, Jianhang and Ding, Yifan and Wu, Wen and Chen, Hao and Gu, Renshu},
  journal={arXiv preprint arXiv:2605.23969},
  year={2026},
  url={https://arxiv.org/abs/2605.23969}
}

@article{fan2026optiselect,
  title={{OptiSelect}: How does the Optimizer Shape Data Curriculum?},
  author={Fan, Simin and Abdollahpoorrostam, Alireza and Jaggi, Martin},
  journal={arXiv preprint arXiv:2610.03432},
  year={2026},
  url={https://arxiv.org/abs/2610.03432}
}

@article{kim2026form,
  title={Form Over Content In Gradient-Based Data Attribution Methods},
  author={Kim, Sunwoo and Jung, Seokwon and Kim, Sohyung and Oh, Seong Joon and Oh, Alice},
  journal={arXiv preprint arXiv:2609.19589},
  year={2026},
  url={https://arxiv.org/abs/2609.19589}
}

@article{liu2026dattri,
  title={{dattri-LLM}: A Unified and Efficient Library for Training Data Attribution at {LLM} Scale},
  author={Liu, Shixuan and Zhou, Tongli and Deng, Junwei and Hu, Pingbang and Ma, Jiaqi W.},
  journal={arXiv preprint arXiv:2609.38767},
  year={2026},
  url={https://arxiv.org/abs/2609.38767}
}

@article{ding2026inrun,
  title={In-Run Data {Shapley} for {Adam} Optimizer},
  author={Ding, Meng and Zhang, Zeqing and Wang, Di and Hu, Lijie},
  journal={arXiv preprint arXiv:2602.00329},
  year={2026},
  url={https://arxiv.org/abs/2602.00329}
}

@article{yuan2026uncertainty,
  title={Uncertainty-Aware Gradient Signal-to-Noise Data Selection for Instruction Tuning},
  author={Yuan, Zhihang and Yue, Chengyu and Huang, Long and Ou, Litu and Shi, Lei},
  journal={arXiv preprint arXiv:2601.13697},
  year={2026},
  url={https://arxiv.org/abs/2601.13697}
}

@article{wang2026drift,
  title={{DRIFT}: Data Selection for {LLM} Instruction Tuning via On-Policy Attribution},
  author={Wang, Zefan and Li, Lincheng and Yu, Tianyu and Yao, Yuan},
  journal={arXiv preprint arXiv:2606.18307},
  year={2026},
  url={https://arxiv.org/abs/2606.18307}
}

@article{zeng2026online,
  title={Online Data Selection Is Implicit Alignment},
  author={Zeng, Aoxiong and Yang, Yuxin and Yang, Xiangquan},
  journal={arXiv preprint arXiv:2607.07023},
  year={2026},
  url={https://arxiv.org/abs/2607.07023}
}

@article{yu2026sign,
  title={Sign-Based Optimizers Are Effective Under Heavy-Tailed Noise},
  author={Yu, Dingzhi and Tao, Hongyi and Wan, Yuanyu and Luo, Luo and Zhang, Lijun},
  journal={arXiv preprint arXiv:2602.07425},
  year={2026},
  url={https://arxiv.org/abs/2602.07425}
}

@article{necula2026highquality,
  title={"High-Quality" Data for Early-Stage Instruction Tuning: An Empirical Study of Data Quality Metrics},
  author={Necula, Vladimir and Gvadzabia, David and Serrano, Sofia},
  journal={arXiv preprint arXiv:2610.33969},
  year={2026},
  url={https://arxiv.org/abs/2610.33969}
}

@article{qin2026hidden,
  title={{Hidden} {Cartographer}: {Hidden} State Guided Data Selection for Dual Budgeted Instruction Tuning},
  author={Qin, Zixuan and Qin, Qi and Li, Erbo and Wei, Ting and Sun, Yifan and Fan, Zhaoxin},
  journal={arXiv preprint arXiv:2610.25215},
  year={2026},
  url={https://arxiv.org/abs/2610.25215}
}

@article{sun2026rethinking,
  title={Rethinking Data Selection as Distribution Matching on Semantic Manifolds},
  author={Sun, Peng and Yang, Yi and Chen, Mingshuo and Li, Chunxiao and Zhu, Yao and Shen, Huawen and Fu, Tianfan},
  journal={arXiv preprint arXiv:2611.08495},
  year={2026},
  url={https://arxiv.org/abs/2611.08495}
}

@article{cui2026learning,
  title={Learning-Zone Energy: Online Data Selection for Efficient {RL} Post-Training},
  author={Cui, Peng and Yang, Boyao and Zhu, Jun},
  journal={arXiv preprint arXiv:2605.17003},
  year={2026},
  url={https://arxiv.org/abs/2605.17003}
}

@inproceedings{han2026what,
  title={What Makes Good Instruction-Tuning Data? An In-Context Learning Perspective},
  author={Han, Guangzeng and Huang, Xiaolei},
  booktitle={Proceedings of the 64th Annual Meeting of the Association for Computational Linguistics (Volume 1: Long Papers)},
  year={2026},
  url={https://arxiv.org/abs/2604.25132}
}

@inproceedings{wang2024greats,
  title={{GREATS}: Online Selection of High-Quality Data for {LLM} Training in Every Iteration},
  author={Wang, Jiachen T. and Wu, Tong and Song, Dawn and Mittal, Prateek and Jia, Ruoxi},
  booktitle={Advances in Neural Information Processing Systems},
  volume={37},
  pages={131197--131223},
  year={2024}
}

@inproceedings{loshchilov2019decoupled,
  title={Decoupled Weight Decay Regularization},
  author={Loshchilov, Ilya and Hutter, Frank},
  booktitle={International Conference on Learning Representations},
  year={2019}
}

@article{nemhauser1978analysis,
  title={An analysis of approximations for maximizing submodular set functions---{I}},
  author={Nemhauser, George L. and Wolsey, Laurence A. and Fisher, Marshall L.},
  journal={Mathematical Programming},
  volume={14},
  number={1},
  pages={265--294},
  year={1978},
  doi={10.1007/BF01588971}
}

@incollection{cornuejols1977uncapacitated,
  title={On the Uncapacitated Location Problem},
  author={Cornuejols, Gerard and Fisher, Marshall and Nemhauser, George L.},
  booktitle={Annals of Discrete Mathematics},
  volume={1},
  pages={163--177},
  publisher={Elsevier},
  year={1977},
  doi={10.1016/S0167-5060(08)70732-5}
}

@article{ghadimi2013stochastic,
  title={Stochastic First- and Zeroth-Order Methods for Nonconvex Stochastic Programming},
  author={Ghadimi, Saeed and Lan, Guanghui},
  journal={SIAM Journal on Optimization},
  volume={23},
  number={4},
  pages={2341--2368},
  year={2013},
  doi={10.1137/120880811}
}

@article{bottou2018optimization,
  title={Optimization Methods for Large-Scale Machine Learning},
  author={Bottou, L{\'e}on and Curtis, Frank E. and Nocedal, Jorge},
  journal={SIAM Review},
  volume={60},
  number={2},
  pages={223--311},
  year={2018},
  doi={10.1137/16M1080173}
}

@inproceedings{karimi2016linear,
  title={Linear Convergence of Gradient and Proximal-Gradient Methods Under the {Polyak--{\L}ojasiewicz} Condition},
  author={Karimi, Hamed and Nutini, Julie and Schmidt, Mark},
  booktitle={Joint European Conference on Machine Learning and Knowledge Discovery in Databases},
  pages={795--811},
  publisher={Springer},
  year={2016},
  doi={10.1007/978-3-319-46128-1_50}
}

@book{bauschke2017convex,
  title={Convex Analysis and Monotone Operator Theory in {Hilbert} Spaces},
  author={Bauschke, Heinz H. and Combettes, Patrick L.},
  edition={2nd},
  publisher={Springer},
  year={2017},
  doi={10.1007/978-3-319-48311-5}
}

@inproceedings{peng2024energy,
  title={Energy-based Automated Model Evaluation},
  author={Peng, Ru and Zou, Heming and Wang, Haobo and Zeng, Yawen and Huang, Zenan and Zhao, Junbo},
  booktitle={International Conference on Learning Representations},
  year={2024},
  url={https://arxiv.org/abs/2401.12689}
}

@inproceedings{zou2025flylora,
  title={{FlyLoRA}: Boosting Task Decoupling and Parameter Efficiency via Implicit Rank-Wise Mixture-of-Experts},
  author={Zou, Heming and Zang, Yunliang and Xu, Wutong and Zhu, Yao and Ji, Xiangyang},
  booktitle={Advances in Neural Information Processing Systems},
  volume={38},
  pages={10386--10419},
  year={2025},
  url={https://arxiv.org/abs/2510.08396}
}

@article{zhang2026coda,
  title={Co-Adaptive Multi-Task {LoRA}: Transfer-Aware, Label-Free Control of Domain Participation},
  author={Zhang, Wei and Tang, Lin and Zhao, Ming and Wang, Yuxuan},
  journal={arXiv preprint arXiv:2607.03522},
  year={2026},
  url={https://arxiv.org/abs/2607.03522}
}

@article{tang2026merge,
  title={Predicting Mergeability of Parameter-Efficient Fine-Tuning Updates},
  author={Tang, Lin and Zhang, Wei and Li, Jing and Chen, Hongyu and Zhao, Ming and Wang, Yuxuan},
  journal={arXiv preprint arXiv:2606.19549},
  year={2026},
  url={https://arxiv.org/abs/2606.19549}
}

\appendix
\section{Algorithms and Factored Ghost Implementation}
\label{app:alg}

\subsection{Stochastic-Greedy Variant (\method{}-SG)}
\label{app:sg}

For larger candidate pool sizes ($b \ge 64$), evaluating all $b - |\mathcal{S}|$ unselected candidates at every greedy step incurs $O(k b^2)$ scalar operations. Algorithm~\ref{alg:prism_sg} accelerates selection by replacing exhaustive greedy scanning with stochastic greedy sampling~\citep{mirzasoleiman2015lazier}. Given an accuracy tolerance $\varepsilon \in (0,1)$, each step samples uniformly at random a subset $\mathcal{R}_r \subseteq [b]\setminus\mathcal{S}$ of size $q = \lceil (b/k)\log(1/\varepsilon)\rceil$ and maximizes the conic marginal gain over $\mathcal{R}_r$. Because $F_t^{\mathrm{cone}}$ is normalized, nonnegative, monotone, and submodular, \method{}-SG reduces selection overhead to $O(b^2 \log(1/\varepsilon))$ while retaining a $(1-1/e-\varepsilon)$ expected approximation guarantee (Proposition~\ref{prop:sg_bound}).

\begin{algorithm}[t]
\footnotesize
\DontPrintSemicolon
\SetAlgoLined
\caption{\method{}-SG: Stochastic-Greedy Conic Selection}
\label{alg:prism_sg}
\KwInput{parameters $w_t$; batch $\mathcal{B}_t=\{z_i\}_{i=1}^b$; size $k$; tolerance $\varepsilon\in(0,1)$; metric $M_t$}
Compute $(K_t)_{ij}$ via~\eqref{eq:precond_ghost} and $s_{t,ij}^{\mathrm{cone}} \leftarrow [\max(0,(K_t)_{ij})]^2/((K_t)_{jj}+\delta)$\;
Set $q \leftarrow \lceil (b/k)\log(1/\varepsilon)\rceil$, $\mathcal{S}\leftarrow\emptyset$, $u_i \leftarrow 0$ for $i\in[b]$\;
\For{$r = 1, \dots, k$}{
  Sample $\mathcal{R}_r \subseteq [b]\setminus\mathcal{S}$ uniformly with $|\mathcal{R}_r|=\min\{q,\,b-|\mathcal{S}|\}$\;
  $j^\star \leftarrow \arg\max_{j\in\mathcal{R}_r} \sum_{i=1}^b \max(0,\, s_{t,ij}^{\mathrm{cone}} - u_i)$\;
  $\mathcal{S}\leftarrow\mathcal{S}\cup\{j^\star\}$;\quad $u_i \leftarrow \max(u_i,\, s_{t,ij^\star}^{\mathrm{cone}})\ \forall i\in[b]$\;
}
Assign $\sigma(i) \leftarrow \arg\max_{j\in\mathcal{S}} s_{t,ij}^{\mathrm{cone}}$ and $C_j \leftarrow \{i:\sigma(i)=j\}$\;
Compute $\gamma_{t,j} \leftarrow \sum_{i\in C_j} \max(0,(K_t)_{ij})/((K_t)_{jj}+\delta)$\;
Obtain $\hat{g}_t = \nabla_w \big[\frac{1}{b}\sum_{j\in\mathcal{S}} \gamma_{t,j}\ell(w_t,z_j)\big]$ via one backward pass on $\mathcal{S}$\;
Update $w_{t+1}$ and factored metric $M_{t+1}$\;
\end{algorithm}

\subsection{Factored Ghost Inner Products for LoRA and Linear Modules}
\label{app:impl}

Consider a frozen linear backbone matrix $W_0 \in \R^{d_{\mathrm{out}}\times d_{\mathrm{in}}}$ augmented with LoRA factors $A \in \R^{r\times d_{\mathrm{in}}}$ and $B \in \R^{d_{\mathrm{out}}\times r}$~\citep{hu2022lora}:
\begin{equation}
y_i = W_0 x_i + \frac{\alpha}{r} B A x_i, \qquad h_i = A x_i \in \R^r.
\end{equation}
For sequence inputs $x_i \in \R^{L_i \times d_{\mathrm{in}}}$ (or summed over non-pad tokens $m \in [L_i]$), let $\theta_{i,m} = \nabla_{y_{i,m}} \ell_i \in \R^{d_{\mathrm{out}}}$ and $\psi_{i,m} = \frac{\alpha}{r} B^\top \theta_{i,m} \in \R^r$. The per-example LoRA gradients are
\begin{equation}
\nabla_A \ell_i = \sum_{m=1}^{L_i} \psi_{i,m} x_{i,m}^\top, \qquad
\nabla_B \ell_i = \frac{\alpha}{r}\sum_{m=1}^{L_i} \theta_{i,m} h_{i,m}^\top.
\end{equation}
Under the factored metric $M_A = D_{\psi} \otimes D_x$ and $M_B = D_{\theta} \otimes D_h$, we rescale the lightweight token activations and backpropagated signals prior to contraction:
\begin{equation}
\begin{aligned}
\tilde{x}_{i,m} &= D_x^{1/2} x_{i,m}, & \tilde{\psi}_{i,m} &= D_\psi^{1/2} \psi_{i,m}, \\
\tilde{\theta}_{i,m} &= D_\theta^{1/2} \theta_{i,m}, & \tilde{h}_{i,m} &= D_h^{1/2} h_{i,m}.
\end{aligned}
\end{equation}
Because $(D_\psi \otimes D_x)$ is separable across row and column indices, the preconditioned Frobenius inner product decomposes identically to~\eqref{eq:ghost}:
\begin{equation}
\begin{aligned}
&\ip{\nabla_A \ell_i}{\nabla_A \ell_j}_{M_A} \\
&\qquad= \sum_{m=1}^{L_i}\sum_{m'=1}^{L_j} \ip{\tilde{\psi}_{i,m}}{\tilde{\psi}_{j,m'}}\,\ip{\tilde{x}_{i,m}}{\tilde{x}_{j,m'}},
\end{aligned}
\end{equation}
and analogously for $\nabla_B \ell_i$. Following~\citet{wang2024greats}, contracting first along the low-rank dimension $r$ or sequence dimension via batched matrix multiplication avoids materializing any $d_{\mathrm{out}}\times d_{\mathrm{in}}$ or $r\times d_{\mathrm{in}}$ gradient tensor.

\paragraph{State and complexity accounting.}
Table~\ref{tab:complexity} compares asymptotic memory and time per step. Whereas naive gradient matching materializes $O(bp)$ floats, \method{} maintains only $O(\sum_\ell (d_{\mathrm{in},\ell}+d_{\mathrm{out},\ell}))$ metric vectors and an $O(b^2)$ Gram matrix.

\begin{table}[t]
\centering
\footnotesize
\setlength{\tabcolsep}{2.4pt}
\begin{tabular}{l ccc}
\toprule
\textbf{Method} & \textbf{Grad Mem} & \textbf{Metric} & \textbf{Sel.\ Time} \\
\midrule
Naive GradMatch & $O(bp)$ & $O(1)$ & $O(k b p)$ \\
Euclidean-FL & $O(b L d)$ & $O(1)$ & $O(b^2 d + k b^2)$ \\
ZeroOrder (ZO) & $O(b)$ & $O(1)$ & $O(b d_{\mathrm{p}} + k b^2)$ \\
\textbf{\method{}} & $O(b L d)$ & $O(\sum_\ell d_\ell)$ & $O(b^2 d + k b^2)$ \\
\textbf{\method{}-SG} & $O(b L d)$ & $O(\sum_\ell d_\ell)$ & $O(b^2 d + b^2 \log \tfrac{1}{\varepsilon})$ \\
\bottomrule
\end{tabular}
\caption{Per-step memory and complexity ($p$: parameters, $d$: width, $L$: sequence length, $d_{\mathrm{p}}$: ZO projection dim, $b$: pool, $k$: subset).}
\label{tab:complexity}
\end{table}

\section{Proofs for Section~\ref{sec:theory}}
\label{app:proofs}

\subsection{Proof of Proposition~\ref{prop:submod}}
\label{app:proof_submod}

\begin{proof}
By definition~\eqref{eq:conic_fl}, $F_t^{\mathrm{cone}}(\emptyset) = 0$, and since $s_{t,ij}^{\mathrm{cone}} \ge 0$ for all $i,j \in [b]$, $F_t^{\mathrm{cone}}(\mathcal{S}) \ge 0$ for every $\mathcal{S} \subseteq \mathcal{B}_t$. Let $\mathcal{S} \subseteq \mathcal{T} \subseteq \mathcal{B}_t$ and $z_u \in \mathcal{B}_t \setminus \mathcal{T}$. For each candidate $i \in [b]$, define $f_i(\mathcal{S}) := \max_{j \in \mathcal{S}} s_{t,ij}^{\mathrm{cone}}$ (with $f_i(\emptyset)=0$). Clearly $f_i(\mathcal{S} \cup \{z_u\}) - f_i(\mathcal{S}) = \max(0, s_{t,iu}^{\mathrm{cone}} - f_i(\mathcal{S})) \ge 0$, proving monotonicity. Moreover, because $\mathcal{S} \subseteq \mathcal{T}$ implies $f_i(\mathcal{S}) \le f_i(\mathcal{T})$, we have
\begin{equation}
\max\big(0,\, s_{t,iu}^{\mathrm{cone}} - f_i(\mathcal{S})\big)
\ge \max\big(0,\, s_{t,iu}^{\mathrm{cone}} - f_i(\mathcal{T})\big),
\end{equation}
so $f_i$ is submodular for each $i$~\citep{cornuejols1977uncapacitated}. Since a nonnegative sum of monotone submodular functions is monotone submodular, $F_t^{\mathrm{cone}}(\mathcal{S}) = \sum_{i=1}^b f_i(\mathcal{S})$ is normalized, monotone, and submodular. By the classical theorem of~\citet{nemhauser1978analysis}, greedy maximization under $|\mathcal{S}| \le k$ returns $\mathcal{S}_G$ satisfying
\begin{equation}
F_t^{\mathrm{cone}}(\mathcal{S}_G) \ge \left(1 - \frac{1}{e}\right) F_t^{\mathrm{cone}}(\mathcal{S}^\star).
\end{equation}
Using~\eqref{eq:conic_cost}, $c_t^{\mathrm{cone}}(\mathcal{S}) = E_t - F_t^{\mathrm{cone}}(\mathcal{S})$ for all nonempty $\mathcal{S} \subseteq \mathcal{B}_t$. Substituting $F_t^{\mathrm{cone}}(\mathcal{S}) = E_t - c_t^{\mathrm{cone}}(\mathcal{S})$ for $\mathcal{S}_G$ and $\mathcal{S}^\star$ yields
\begin{align}
E_t - c_t^{\mathrm{cone}}(\mathcal{S}_G) &\ge \left(1 - \frac{1}{e}\right)\big(E_t - c_t^{\mathrm{cone}}(\mathcal{S}^\star)\big) \nonumber\\
\implies c_t^{\mathrm{cone}}(\mathcal{S}_G) &\le \frac{1}{e}E_t + \left(1 - \frac{1}{e}\right)c_t^{\mathrm{cone}}(\mathcal{S}^\star),
\end{align}
which establishes~\eqref{eq:energy_bound}.
\end{proof}

\subsection{Proof of Proposition~\ref{prop:radial_bound}}
\label{app:proof_radial}

\begin{proof}
For each candidate $i \in [b]$, let $\sigma(i) = \sigma_{\mathcal{S}_t}(i) \in \arg\max_{j\in\mathcal{S}_t} s_{t,ij}^{\mathrm{cone}}$ and define the nonnegative radial projection of $\tilde{g}_{t,i} = M_t^{1/2} g_{t,i}$ onto the ray generated by $\tilde{g}_{t,\sigma(i)}$ as
\begin{equation}
p_i := \alpha_i \tilde{g}_{t,\sigma(i)}, \qquad \alpha_i := \frac{\max(0,\, (K_t)_{i,\sigma(i)})}{(K_t)_{\sigma(i),\sigma(i)} + \delta}.
\end{equation}
Because the cells $\{C_j : j \in \mathcal{S}_t\}$ partition $[b]$, regrouping the sum over $i \in [b]$ by $j \in \mathcal{S}_t$ gives
\begin{equation}
\begin{aligned}
\frac{1}{b}\sum_{i=1}^b p_i &= \frac{1}{b}\sum_{j\in\mathcal{S}_t}\left(\sum_{i\in C_j}\alpha_i\right)\tilde{g}_{t,j} \\
&= \frac{1}{b}\sum_{j\in\mathcal{S}_t}\gamma_{t,j}\tilde{g}_{t,j} = M_t^{1/2}\hat{g}_t.
\end{aligned}
\end{equation}
Therefore, by the triangle inequality and Jensen's inequality ($\big(\frac{1}{b}\sum_{i=1}^b a_i\big)^2 \le \frac{1}{b}\sum_{i=1}^b a_i^2$),
\begin{equation}
\begin{aligned}
\nrmM{g_t - \hat{g}_t}^2
&= \left\|\frac{1}{b}\sum_{i=1}^b (\tilde{g}_{t,i} - p_i)\right\|_2^2 \\
&\le \left(\frac{1}{b}\sum_{i=1}^b \|\tilde{g}_{t,i} - p_i\|_2\right)^2
\le \frac{1}{b}\sum_{i=1}^b \|\tilde{g}_{t,i} - p_i\|_2^2.
\end{aligned}
\end{equation}
When $(K_t)_{i,\sigma(i)} \le 0$, $p_i = 0$ and therefore $\|\tilde{g}_{t,i} - p_i\|_2^2 = (K_t)_{ii} - s_{t,i,\sigma(i)}^{\mathrm{cone}}$ (since $s_{t,i,\sigma(i)}^{\mathrm{cone}} = 0$). When $(K_t)_{i,\sigma(i)} > 0$, expanding the squared norm yields
\begin{equation}
\begin{aligned}
\|\tilde{g}_{t,i} - p_i\|_2^2
&= (K_t)_{ii} - 2\alpha_i (K_t)_{i,\sigma(i)} + \alpha_i^2 (K_t)_{\sigma(i),\sigma(i)} \\
&\le (K_t)_{ii} - \frac{(K_t)_{i,\sigma(i)}^2}{(K_t)_{\sigma(i),\sigma(i)} + \delta} \\
&= (K_t)_{ii} - s_{t,i,\sigma(i)}^{\mathrm{cone}},
\end{aligned}
\end{equation}
because $(K_t)_{\sigma(i),\sigma(i)} \le (K_t)_{\sigma(i),\sigma(i)}{+}\delta$ (exact equality as $\delta \to 0^+$). Summing over $i \in [b]$ proves $\nrmM{g_t - \hat{g}_t}^2 \le \frac{1}{b}c_t^{\mathrm{cone}}(\mathcal{S}_t)$, establishing part (i). Part (ii) follows directly from~\eqref{eq:radial_weights}: if $\ipM{g_{t,i}}{g_{t,j}} = (K_t)_{ij} \le 0$ for all $j\in\mathcal{S}_t$, then $\alpha_i = 0$, contributing zero to $\gamma_{t,\sigma(i)}$, whereas Euclidean assignment increments $|C_{\sigma(i)}|$ by $1$.
\end{proof}

\subsection{Proof of Theorem~\ref{thm:nonconvex} and Corollary~\ref{cor:prism_conv}}
\label{app:proof_conv}

We first prove a preconditioned one-step descent lemma analogous to~\citet{ghadimi2013stochastic} and~\citet{bottou2018optimization}.

\begin{lemma}[Preconditioned one-step descent]
\label{lem:descent}
Under Assumption~\ref{asm:smooth}, if $w_{t+1} = w_t - \eta_t M_t \hat{g}_t$ with $\mu_M I \preceq M_t \preceq L_M I$ and $0 < \eta_t \le 1/(4 L L_M)$, then
\begin{equation}
\begin{aligned}
\E_t[f(w_{t+1})] &\le f(w_t) - \frac{\eta_t}{2}\|\nabla f(w_t)\|_{M_t}^2 \\
&\quad + \frac{L L_M \eta_t^2 \sigma_M^2}{b} + \frac{5\eta_t}{4}\E_t\big[\|e_t\|_{M_t}^2\big].
\end{aligned}
\label{eq:one_step}
\end{equation}
\end{lemma}

\begin{proof}
By $L$-smoothness of $f$,
\begin{equation}
\begin{aligned}
f(w_{t+1}) &\le f(w_t) - \eta_t\ip{\nabla f(w_t)}{M_t \hat{g}_t} \\
&\quad + \frac{L\eta_t^2}{2}\|M_t \hat{g}_t\|_2^2.
\end{aligned}
\end{equation}
Since $M_t \preceq L_M I$, $\|M_t \hat{g}_t\|_2^2 = \hat{g}_t^\top M_t^2 \hat{g}_t \le L_M \|\hat{g}_t\|_{M_t}^2$. Writing $\hat{g}_t = g_t + e_t$ and taking conditional expectation $\E_t[\cdot]$ (with $w_t, M_t$ fixed and $\E_t[g_t] = \nabla f(w_t)$):
\begin{equation}
\begin{aligned}
\E_t[f(w_{t+1})] &\le f(w_t) - \eta_t \|\nabla f(w_t)\|_{M_t}^2 \\
&\quad - \eta_t \E_t\ipM{\nabla f(w_t)}{e_t} \\
&\quad + \frac{L L_M \eta_t^2}{2}\E_t\big[\|g_t + e_t\|_{M_t}^2\big].
\end{aligned}
\end{equation}
Applying Cauchy--Schwarz and Young's inequality gives $-\ipM{\nabla f(w_t)}{e_t} \le \frac{1}{4}\|\nabla f(w_t)\|_{M_t}^2 + \|e_t\|_{M_t}^2$, while $\|g_t + e_t\|_{M_t}^2 \le 2\|g_t\|_{M_t}^2 + 2\|e_t\|_{M_t}^2$. By Assumption~\ref{asm:smooth},
\begin{equation*}
\begin{aligned}
\E_t\|g_t\|_{M_t}^2 &= \|\nabla f(w_t)\|_{M_t}^2 + \E_t\|g_t - \nabla f(w_t)\|_{M_t}^2 \\
&\le \|\nabla f(w_t)\|_{M_t}^2 + \frac{\sigma_M^2}{b}.
\end{aligned}
\end{equation*}
Substituting these inequalities yields
\begin{equation}
\begin{aligned}
\E_t[f(w_{t+1})] &\le f(w_t) \\
&\quad - \left(\frac{3\eta_t}{4} - L L_M \eta_t^2\right)\|\nabla f(w_t)\|_{M_t}^2 \\
&\quad + \frac{L L_M \eta_t^2 \sigma_M^2}{b} + \big(\eta_t + L L_M \eta_t^2\big)\E_t\|e_t\|_{M_t}^2.
\end{aligned}
\end{equation}
Using $\eta_t \le 1/(4 L L_M)$ gives $\frac{3\eta_t}{4} - L L_M \eta_t^2 \ge \frac{\eta_t}{2}$ and $\eta_t + L L_M \eta_t^2 \le \frac{5\eta_t}{4}$, proving~\eqref{eq:one_step}.
\end{proof}

\begin{proof}[Proof of Theorem~\ref{thm:nonconvex}]
Taking total expectation over Lemma~\ref{lem:descent} and summing over $t=0,\dots,T-1$ produces a telescoping sum on the objective differences:
\begin{equation}
\begin{aligned}
\frac{1}{2}\sum_{t=0}^{T-1}\eta_t \E\|\nabla f(w_t)\|_{M_t}^2
&\le f(w_0) - \E[f(w_T)] \\
&\quad + \frac{L L_M \sigma_M^2}{b}\sum_{t=0}^{T-1}\eta_t^2 \\
&\quad + \frac{5}{4}\sum_{t=0}^{T-1}\eta_t \E\|e_t\|_{M_t}^2.
\end{aligned}
\end{equation}
Since $f(w_T) \ge f^\star$, $f(w_0) - \E[f(w_T)] \le \Delta_f$. Dividing both sides by $\frac{1}{2}\sum_{t=0}^{T-1}\eta_t$ yields~\eqref{eq:thm_conv}.
\end{proof}

\begin{proof}[Proof of Corollary~\ref{cor:prism_conv}]
By Proposition~\ref{prop:radial_bound}, $\|e_t\|_{M_t}^2 = \nrmM{\hat{g}_t - g_t}^2 \le c_t^{\mathrm{cone}}(\mathcal{S}_t)/b$ pointwise for every batch realization. Substituting this bound and constant $\eta_t = \eta$ into~\eqref{eq:thm_conv} gives
\begin{equation}
\begin{aligned}
\frac{1}{T}\sum_{t=0}^{T-1}\E\|\nabla f(w_t)\|_{M_t}^2 &\le \frac{2\Delta_f}{\eta T} + \frac{2 L L_M \eta \sigma_M^2}{b} \\
&\quad + \frac{5}{2}\delta_T^2.
\end{aligned}
\end{equation}
Choosing $\eta = \sqrt{\Delta_f b / (L L_M \sigma_M^2 T)}$ (valid for $T \ge 16 L L_M \Delta_f b / \sigma_M^2$ so that $\eta \le 1/(4 L L_M)$) equalizes the first two terms to $4\sqrt{\Delta_f L L_M \sigma_M^2 / (bT)}$, proving~\eqref{eq:cor_rate}.
\end{proof}

\section{Additional Theoretical Results}
\label{app:extra_theory}

\subsection{Linear Convergence under the Polyak--{\L}ojasiewicz Condition}
\label{app:pl}

\begin{assumption}[Preconditioned PL condition]
\label{asm:pl}
There exists $\mu > 0$ such that for all $w\in\R^p$, $\|\nabla f(w)\|_{M_t}^2 \ge 2\mu\big(f(w) - f^\star\big)$~\citep{karimi2016linear}. (Since $M_t \succeq \mu_M I$, the standard Euclidean PL constant $\mu_{\mathrm{euc}}$ implies $\mu \ge \mu_M \mu_{\mathrm{euc}}$.)
\end{assumption}

\begin{theorem}[PL convergence of \method{}]
\label{thm:pl}
Suppose Assumptions~\ref{asm:smooth} and~\ref{asm:pl} hold, and \method{} runs with constant step size $0 < \eta \le \min\{\frac{1}{4 L L_M}, \frac{1}{\mu}\}$ and bounded expected conic cost $\frac{1}{b}\E[c_t^{\mathrm{cone}}(\mathcal{S}_t)] \le \delta^2$ for all $t$. Then
\begin{equation}
\begin{aligned}
&\E[f(w_T) - f^\star] \le (1 - \mu\eta)^T \Delta_f \\
&\qquad + \big(1 - (1-\mu\eta)^T\big)\left(\frac{L L_M \eta \sigma_M^2}{\mu b} + \frac{5\delta^2}{4\mu}\right).
\end{aligned}
\end{equation}
\end{theorem}

\begin{proof}
Combining Lemma~\ref{lem:descent} with Assumption~\ref{asm:pl} ($-\frac{\eta}{2}\|\nabla f(w_t)\|_{M_t}^2 \le -\mu\eta(f(w_t)-f^\star)$) and Proposition~\ref{prop:radial_bound} ($\E_t\|e_t\|_{M_t}^2 \le \frac{1}{b}\E_t[c_t^{\mathrm{cone}}(\mathcal{S}_t)]$) and taking total expectation gives
\begin{equation}
\begin{aligned}
\E[f(w_{t+1}) - f^\star] &\le (1-\mu\eta)\E[f(w_t)-f^\star] \\
&\quad + \frac{L L_M \eta^2 \sigma_M^2}{b} + \frac{5\eta}{4}\delta^2.
\end{aligned}
\end{equation}
Unrolling the linear contraction $q = 1-\mu\eta \in [0,1)$ over $T$ steps and summing the geometric series $\sum_{r=0}^{T-1} q^r = (1 - q^T)/(\mu\eta)$ yields the stated bound.
\end{proof}

\subsection{Asymptotic Stationarity under Diminishing Step Sizes}
\label{app:diminishing}

\begin{corollary}[Asymptotic stationarity]
\label{cor:diminishing}
Under Theorem~\ref{thm:nonconvex}, suppose $\sum_{t=0}^\infty \eta_t = \infty$, $\sum_{t=0}^\infty \eta_t^2 < \infty$, and $\sum_{t=0}^{T-1}\eta_t \E[c_t^{\mathrm{cone}}(\mathcal{S}_t)/b] \big/ \sum_{t=0}^{T-1}\eta_t \to 0$ as $T\to\infty$. Then
\begin{equation}
\frac{\sum_{t=0}^{T-1}\eta_t \E\|\nabla f(w_t)\|_{M_t}^2}{\sum_{t=0}^{T-1}\eta_t} \to 0.
\end{equation}
\end{corollary}

\begin{proof}
Immediate by taking $T\to\infty$ in~\eqref{eq:thm_conv} and using the divergence of $\sum_t \eta_t$ against finite $\Delta_f$ and $\sum_t \eta_t^2$.
\end{proof}

\subsection{Stochastic-Greedy Conic Cost Bound}
\label{app:sg_theory}

\begin{proposition}[Stochastic-greedy conic cost]
\label{prop:sg_bound}
Let $\mathcal{S}_{\mathrm{SG}}$ be returned by Algorithm~\ref{alg:prism_sg} with parameter $\varepsilon \in (0,1)$ and let $\mathcal{S}^\star$ be an optimal size-$k$ subset. Then
\begin{equation}
\begin{aligned}
\E_{\mathrm{SG}}\big[c_t^{\mathrm{cone}}(\mathcal{S}_{\mathrm{SG}})\big]
&\le \left(\frac{1}{e} + \varepsilon\right) E_t \\
&\quad + \left(1 - \frac{1}{e} - \varepsilon\right) c_t^{\mathrm{cone}}(\mathcal{S}^\star).
\end{aligned}
\end{equation}
\end{proposition}

\begin{proof}
Because $F_t^{\mathrm{cone}}$ is normalized, nonnegative, monotone, and submodular (Proposition~\ref{prop:submod}), the stochastic-greedy theorem of~\citet{mirzasoleiman2015lazier} gives $\E_{\mathrm{SG}}[F_t^{\mathrm{cone}}(\mathcal{S}_{\mathrm{SG}})] \ge (1 - 1/e - \varepsilon) F_t^{\mathrm{cone}}(\mathcal{S}^\star)$. Substituting $F_t^{\mathrm{cone}}(\mathcal{S}) = E_t - c_t^{\mathrm{cone}}(\mathcal{S})$ completes the proof.
\end{proof}

\subsection{Non-Submodularity of Second-Order Validation Surrogates}
\label{app:nonsubmod}

\begin{proposition}[Counterexample to validation-utility submodularity]
\label{prop:nonsubmod}
Consider the second-order Taylor validation-utility surrogate
\begin{equation*}
\begin{aligned}
\tilde{U}_t(\mathcal{S}) &= \eta_t \ip{g_t^{\mathrm{val}}}{\sum_{j\in\mathcal{S}} g_{t,j}} \\
&\quad - \frac{\eta_t^2}{2}\Big(\sum_{j\in\mathcal{S}} g_{t,j}\Big)^\top H_t^{\mathrm{val}}\Big(\sum_{j\in\mathcal{S}} g_{t,j}\Big)
\end{aligned}
\end{equation*}
used in Taylor validation selectors~\citep{wang2024greats}. There exist a smooth loss, a two-example batch $\{z_1, z_2\}$, and a validation point $z_{\mathrm{val}}$ with $H_t^{\mathrm{val}} = h I \succ 0$ for which $\tilde{U}_t$ violates submodularity.
\end{proposition}

\begin{proof}
Let $w \in \R$ at $w_t = 0$, $\ell(w, z_1) = \frac{1}{2}(w+1)^2$, $\ell(w, z_2) = \frac{1}{2}(w-1)^2$, and $\ell(w, z_{\mathrm{val}}) = \frac{h}{2}w^2$ with $h > 0$. Then $g_1 = 1$, $g_2 = -1$, $g^{\mathrm{val}} = 0$, and $H^{\mathrm{val}} = h > 0$. Since $\tilde{U}_t(\emptyset) = 0$, submodularity on $\{z_1,z_2\}$ requires $\tilde{U}_t(\{z_1\}) + \tilde{U}_t(\{z_2\}) \ge \tilde{U}_t(\{z_1,z_2\})$. Expanding yields $\tilde{U}_t(\{z_1\}) + \tilde{U}_t(\{z_2\}) - \tilde{U}_t(\{z_1,z_2\}) = \eta_t^2 g_1 H^{\mathrm{val}} g_2 = -\eta_t^2 h < 0$, violating submodularity. Thus, greedy selection over $\tilde{U}_t$ lacks a structural $(1-1/e)$ guarantee without additional curvature restrictions~\citep{bauschke2017convex}.
\end{proof}

\section{Detailed Experimental Setup}
\label{app:setup}

\begin{table}[t]
\centering
\footnotesize
\setlength{\tabcolsep}{3pt}
\begin{tabularx}{\columnwidth}{@{}l X@{}}
\toprule
\textbf{Hyperparameter} & \textbf{Value} \\
\midrule
LoRA rank $r$ / alpha $\alpha$ & $8$ / $16$ \\
LoRA target modules & q, k, v, o, gate, up, down \\
Candidate pool $b$ / selected $k$ & $16$ / $8$ ($\rho = 0.5$) \\
Factored metric EMA $\beta_2, \epsilon$ & $0.999$, $10^{-8}$ \\
Conic stabilizer $\delta$ & $10^{-12}$ \\
Optimizer & AdamW ($\beta_1{=}0.9$, wd $=0.0$) \\
Learning rate & $2{\times}10^{-5}$ (MMLU/TyDiQA), $5{\times}10^{-5}$ (BBH/GSM8K) \\
LR schedule / warmup & Cosine decay / 3\% linear warmup \\
Main horizon (steps) & 1{,}000 (MMLU), 800 (BBH/GSM8K), 300 (TyDiQA/SAMSum) \\
Precision & bfloat16 \\
Seeds & 3 (2 subjects $\times$ 3 for MMLU) \\
Hardware & $4{\times}$ NVIDIA A100 80GB \\
\bottomrule
\end{tabularx}
\caption{Default training and \method{} hyperparameters across configurations.}
\label{tab:hparams}
\end{table}

\paragraph{Datasets and evaluation protocols.}
For \textbf{LESS $\to$ MMLU}~\citep{xia2024less,hendrycks2021measuring}, we fine-tune Qwen2.5-1.5B on the LESS mixed instruction pool and evaluate answer perplexity and zero-shot accuracy on two held-out MMLU subjects (\texttt{sociology} and \texttt{us\_foreign\_policy}) across 3 seeds ($n{=}6$). For \textbf{Tulu-2 $\to$ BBH}~\citep{ivison2023camels,suzgun2023challenging}, we fine-tune Llama-3.1-8B on Tulu-2 SFT and evaluate 3-shot chain-of-thought exact match on all 27 BBH subtasks. For \textbf{LESS $\to$ TyDiQA}~\citep{clark2020tydi}, we fine-tune Mistral-7B and evaluate macro F1 across the GoldP dev split. For \textbf{Tulu-2 $\to$ GSM8K}~\citep{cobbe2021training}, we fine-tune Qwen2.5-7B and evaluate zero-shot exact-match accuracy on the GSM8K test split (1,319 problems). For the cold-start boundary study, we fine-tune raw Qwen3-8B-Base~\citep{yang2025qwen3} on \textbf{SAMSum}~\citep{gliwa2019samsum} using ChatML formatting and report ROUGE-1/2/L.

\paragraph{Baseline configurations.}
GREATS~\citep{wang2024greats}, OPUS~\citep{wang2026opus}, and Filter-then-Weight~\citep{wang2026filter} all receive 5 clean held-out validation/proxy examples drawn from the target task's validation split; OPUS and Filter-then-Weight use our exact LoRA ghost inner products for controlled parity. Euclidean-FL uses unpreconditioned ghost distances $d_{ij} = \|g_i - g_j\|_2$, diameter shift $s_{ij} = m_t - d_{ij}$, and integer cluster counts $|C_j|$~\citep{mirzasoleiman2020coresets}. ZeroOrder replaces the ghost Gram matrix with a 2-point Rademacher zeroth-order gradient sketch matching~\citet{nguyen2025minibatch} while keeping our conic facility-location rule fixed.

\section{Supplementary Experimental Results}
\label{app:extra_exp}

\subsection{Complementary Metrics}
\label{app:comp_metrics}

Table~\ref{tab:mmlu_acc} reports MMLU classification accuracy alongside the answer perplexity in Table~\ref{tab:main}, and Table~\ref{tab:samsum_full} reports ROUGE-1/2/L on Qwen3-8B-Base / SAMSum at both the 300-step cold-start horizon and the 1{,}000-step extended horizon.

\begin{table}[h]
\centering
\small
\setlength{\tabcolsep}{6pt}
\begin{tabular}{l cc}
\toprule
\textbf{Method} & \textbf{MMLU PPL ($\downarrow$)} & \textbf{MMLU Acc (\% $\uparrow$)} \\
\midrule
Regular & $5.84 \pm 0.41$ & $49.8 \pm 0.9$ \\
Random & $6.42 \pm 0.53$ & $48.1 \pm 1.1$ \\
GREATS & $\mathbf{4.62 \pm 0.18}$ & $\mathbf{52.4 \pm 0.5}$ \\
OPUS & $4.79 \pm 0.21$ & $51.9 \pm 0.6$ \\
Euclidean-FL & $5.61 \pm 0.36$ & $50.5 \pm 0.7$ \\
ZeroOrder & $5.92 \pm 0.44$ & $49.4 \pm 0.8$ \\
\rowcolor{oursbg}
\textbf{\method{}} & $5.14 \pm 0.27$ & $51.6 \pm 0.5$ \\
\bottomrule
\end{tabular}
\caption{Complementary MMLU metrics on Qwen2.5-1.5B (mean $\pm$ std over 2 subjects $\times$ 3 seeds, $n{=}6$).}
\label{tab:mmlu_acc}
\end{table}

\begin{table}[t]
\centering
\scriptsize
\setlength{\tabcolsep}{2.2pt}
\begin{tabular}{l ccc ccc}
\toprule
\textbf{Method} & \multicolumn{3}{c}{\textbf{300 steps (Cold-Start)}} & \multicolumn{3}{c}{\textbf{1{,}000 steps (Extended)}} \\
\cmidrule(lr){2-4}\cmidrule(lr){5-7}
 & R-1 & R-2 & R-L & R-1 & R-2 & R-L \\
\midrule
Regular & \textbf{0.214} & \textbf{0.072} & \textbf{0.163} & 0.341 & 0.132 & 0.274 \\
GREATS & 0.172 & 0.054 & 0.131 & 0.369 & 0.149 & 0.298 \\
OPUS & 0.189 & 0.061 & 0.144 & 0.367 & 0.147 & 0.295 \\
Euclidean-FL & 0.131 & 0.039 & 0.098 & 0.358 & 0.141 & 0.288 \\
ZeroOrder & 0.084 & 0.022 & 0.063 & 0.352 & 0.138 & 0.283 \\
\rowcolor{oursbg}
\textbf{\method{}} & 0.148 & 0.045 & 0.112 & \textbf{0.376} & \textbf{0.154} & \textbf{0.304} \\
\bottomrule
\end{tabular}
\caption{SAMSum ROUGE-1/2/L on Qwen3-8B-Base at 300 steps (unseen template cold-start) vs.\ 1{,}000 steps.}
\label{tab:samsum_full}
\end{table}

\subsection{Run-Level Consistency and Additional Baselines}
\label{app:consistency}

Table~\ref{tab:wins} summarizes matched-run win counts across all $15$ main evaluation runs ($6$ MMLU + $3$ BBH + $3$ TyDiQA + $3$ GSM8K), plus the $3$ extended SAMSum runs ($18$ total). \method{} outperforms ZeroOrder in $18/18$ runs and Euclidean-FL in $17/18$ runs. Table~\ref{tab:extra_baselines} compares \method{} against four additional 2025--2026 validation-free and bi-level baselines---CoLM~\citep{nguyen2025minibatch}, UDS~\citep{zou2025utility}, SLAP~\citep{zou2026slap}, and BLADE~\citep{wang2026blade}---on Llama-3.1-8B (BBH) and Qwen2.5-7B (GSM8K).

\begin{table}[t]
\centering
\footnotesize
\setlength{\tabcolsep}{3.0pt}
\begin{tabular}{l ccc}
\toprule
\textbf{Configuration} & \textbf{Runs} & \textbf{vs.\ ZO} & \textbf{vs.\ Eucl.-FL} \\
\midrule
Qwen2.5-1.5B / MMLU & 6 & 6 / 6 & 6 / 6 \\
Llama-3.1-8B / BBH & 3 & 3 / 3 & 3 / 3 \\
Mistral-7B / TyDiQA & 3 & 3 / 3 & 2 / 3 \\
Qwen2.5-7B / GSM8K & 3 & 3 / 3 & 3 / 3 \\
Qwen3-8B / SAMSum (1k) & 3 & 3 / 3 & 3 / 3 \\
\midrule
\textbf{Total} & \textbf{18} & \textbf{18 / 18} & \textbf{17 / 18} \\
\bottomrule
\end{tabular}
\caption{Matched seed-level win rates of \method{} over ZeroOrder and Euclidean-FL.}
\label{tab:wins}
\end{table}

\begin{table}[t]
\centering
\scriptsize
\setlength{\tabcolsep}{2.4pt}
\begin{tabular}{l c cc}
\toprule
\textbf{Method} & \textbf{Ext.} & \textbf{BBH EM ($\uparrow$)} & \textbf{GSM8K Acc ($\uparrow$)} \\
\midrule
CoLM~\citep{nguyen2025minibatch} & No & $55.4 \pm 0.6$ & $76.0 \pm 0.5$ \\
UDS~\citep{zou2025utility} & No & $56.3 \pm 0.5$ & $76.7 \pm 0.4$ \\
SLAP~\citep{zou2026slap} & No & $56.6 \pm 0.5$ & $77.0 \pm 0.5$ \\
BLADE~\citep{wang2026blade} & Ref. & $57.1 \pm 0.4$ & $77.4 \pm 0.4$ \\
\rowcolor{oursbg}
\textbf{\method{}} & No & $\mathbf{57.6 \pm 0.4}$ & $\mathbf{77.9 \pm 0.4}$ \\
\bottomrule
\end{tabular}
\caption{Comparison against recent 2025--2026 baselines on Llama-3.1-8B (BBH) and Qwen2.5-7B (GSM8K).}
\label{tab:extra_baselines}
\end{table}

\subsection{Heavy-Tailed Label-Noise Stress Test}
\label{app:noise}

Table~\ref{tab:noise} evaluates performance when $0\%$, $20\%$, or $40\%$ of training responses in Tulu-2 are replaced with responses from unrelated prompts (Qwen2.5-7B, GSM8K). Because corrupted examples induce misaligned gradients, Euclidean-FL degrades sharply ($-5.4$ points at $40\%$ noise), whereas \method{} drops only $-1.9$ points and beats clean-validation GREATS by $+1.6$ points at $40\%$ noise.

\begin{table}[t]
\centering
\footnotesize
\setlength{\tabcolsep}{3.0pt}
\begin{tabular}{l ccc c}
\toprule
\textbf{Method} & \textbf{0\%} & \textbf{20\%} & \textbf{40\%} & \textbf{$\Delta_{0\to 40}$} \\
\midrule
Regular & 74.6 & 71.2 & 66.8 & $-7.8$ \\
Euclidean-FL & 75.8 & 73.1 & 70.4 & $-5.4$ \\
ZeroOrder & 75.1 & 72.3 & 69.1 & $-6.0$ \\
GREATS (val.) & 76.8 & 75.4 & 74.4 & $-2.4$ \\
\rowcolor{oursbg}
\textbf{\method{}} & \textbf{77.9} & \textbf{76.9} & \textbf{76.0} & $\mathbf{-1.9}$ \\
\bottomrule
\end{tabular}
\caption{GSM8K accuracy (\%) on Qwen2.5-7B under injected response corruption (0\%, 20\%, 40\%).}
\label{tab:noise}
\end{table}

\subsection{Extended-Horizon Robustness Checks}
\label{app:horizon}

\begin{figure*}[t]
    \centering
    \includegraphics[width=\textwidth]{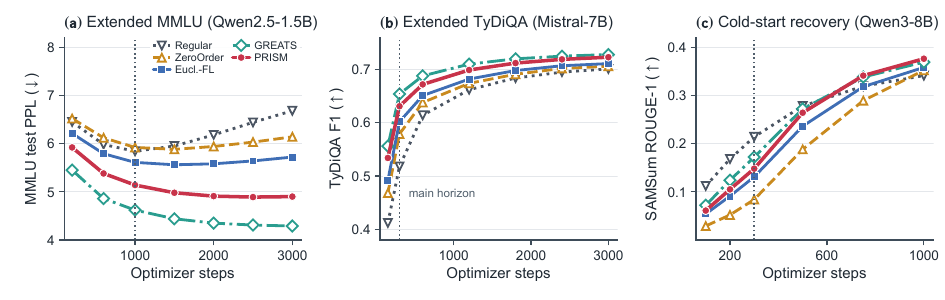}
    \caption{\textbf{Extended-horizon trajectories.} (a) Extending Qwen2.5-1.5B / MMLU from 1{,}000 to 3{,}000 steps preserves the method ordering while Regular begins overfitting. (b) Extending Mistral-7B / TyDiQA to 3{,}000 steps narrows early gaps as optimization saturates, with \method{} remaining the top validation-free selector. (c) Extending Qwen3-8B-Base / SAMSum from 300 to 1{,}000 steps resolves the cold-start deficit: \method{} overtakes Regular around step 700 and finishes highest at $0.376$ ROUGE-1.}
    \label{fig:app_horizons}
\end{figure*}

Figure~\ref{fig:app_horizons} extends training horizons threefold on MMLU (3{,}000 steps), TyDiQA (3{,}000 steps), and SAMSum (1{,}000 steps). On MMLU, Regular begins to overfit past step 1{,}500 ($5.84 \to 6.68$ PPL), while \method{} stabilizes at $4.90$ PPL. On TyDiQA, all methods converge toward $0.70$--$0.73$ F1, with \method{} ($0.723$) closely tracking GREATS ($0.728$). On SAMSum, the 300-step cold-start advantage of Regular completely disappears past step 700 as \method{} finishes at $0.376$ vs.\ $0.341$.

\section{Extended Discussion}
\label{app:discussion}

\paragraph{Why conic geometry outperforms Euclidean distance in LLM batches.}
In Euclidean facility location, covering a candidate $z_i$ by representative $z_j$ incurs penalty $\|g_i - g_j\|_M$, which treats radial magnitude discrepancies and angular misalignment identically. When gradient norms vary by an order of magnitude across sequence lengths or corrupted samples~\citep{zhang2020why,yu2026sign}, two effects harm Euclidean selection: (i) $m_t = \max_{u,v} d_{uv}$ scales with the worst outlier pair, inflating the $(1/e)b m_t$ slack; and (ii) integer counts $|C_j|$ force $z_j$ to absorb full unit mass for misaligned points. Conic projection resolves both by decoupling ray direction from radial amplitude: similarity depends on $\cos_+^2(\tilde{g}_i,\tilde{g}_j)$, and residual error obeys Pythagoras on the positive cone.

\paragraph{Validation-free vs.\ validation-guided selection.}
When a clean, representative target validation set is available and small enough to reflect the evaluation distribution (e.g., MMLU subject validation or TyDiQA dev prompts), GREATS and OPUS can steer training directly toward that target. However, in general instruction tuning (Tulu-2 $\to$ BBH or GSM8K), 5 proxy prompts offer noisy coverage of diverse reasoning patterns, whereas \method{} leverages all $b$ intra-batch gradients as geometric context and outperforms validation-guided methods. Hybridizing \method{} by including a single validation ray inside $\mathcal{S}_t$ is a natural extension when partial supervision exists.

\paragraph{Failure modes and extensions.}
When $b$ is very small ($b \le 8$) or all candidates are nearly orthogonal (early cold-start on an unseen chat format), radial projections shrink most batch energy and slow initial format adaptation; a brief 50--100 step full-batch warmup eliminates this effect. Beyond LoRA SFT, factored preconditioned ghost inner products and conic facility location extend directly to continual fine-tuning, preference optimization, and RL post-training~\citep{cui2026learning,zeng2026online}.

\end{document}